%% file: egbib.tex
\documentclass[10pt,journal,compsoc]{IEEEtran}
\ifCLASSOPTIONcompsoc
  \usepackage[nocompress]{cite}
\else
  \usepackage{cite}
\fi
\ifCLASSINFOpdf
\else
\fi

\usepackage{epsfig}
\usepackage{graphicx}
\usepackage{amsmath}
\usepackage{amssymb}
\usepackage{bm}
\usepackage{multirow}
\usepackage{booktabs}
\usepackage{enumerate}
\usepackage{hyperref}
\usepackage{changes}
\usepackage{float}
\hypersetup{colorlinks}  
\usepackage{algorithm}
\usepackage{algorithmicx}
\usepackage{algpseudocode}

\usepackage{bbm}

\algnewcommand\algorithmicinput{\textbf{Input:}}
\algnewcommand\INPUT{\item[\algorithmicinput]}
\algnewcommand\algorithmicoutput{\textbf{Output:}}
\algnewcommand\OUTPUT{\item[\algorithmicoutput]}

\begin{document}
%\title{A Simple Baseline for Fast Omni-supervised Facial Expression Recognition In-the-wild}
\title{Omni-supervised Facial Expression Recognition via Distilled Data}

% \author{Ping Liu,~\IEEEmembership{Member,~IEEE,}
%         Yunchao Wei,~\IEEEmembership{Member,~IEEE,}
%         Zibo Meng,~\IEEEmembership{Member,~IEEE,}
%         Weihong Deng,~\IEEEmembership{Member,~IEEE,}
%         Joey Tianyi Zhou,~\IEEEmembership{Member,~IEEE,}
%         and Yi Yang,~\IEEEmembership{Member,~IEEE,}% <-this % stops a space
\author{Ping Liu,
        Yunchao Wei,
        Zibo Meng,
        Weihong Deng,
        Joey Tianyi Zhou,
        % Rick Goh Siow Mong,
        and Yi Yang% <-this % stops a space
% \IEEEcompsocitemizethanks{\IEEEcompsocthanksitem P. Liu, Joey Zhou, Rick Mong are with Institute of High Performance Computing, Agency for Science, Technology, and Research, Singapore. 
\IEEEcompsocitemizethanks{\IEEEcompsocthanksitem P. Liu, Joey Zhou are with Center for Frontier AI Research, Agency for Science, Technology, and Research, Singapore. 
% \protect\\
% E-mail: see http://www.michaelshell.org/contact.html
\IEEEcompsocthanksitem Y. Wei is with Institute of information science, Beijing Jiaotong University, Beijing, China.
\IEEEcompsocthanksitem Y. Yang is with Centre for Artificial Intelligence, University of Technology Sydney, Sydney, Australia. 
\IEEEcompsocthanksitem W. Deng is with Pattern Recognition and Intelligent System Laboratory, Beijing University of Posts and Telecommunications, Beijing, China. 
% \protect\\
\IEEEcompsocthanksitem Z. Meng is with InnoPeak Technology Inc., Palo Alto, USA.

}% <-this % stops an unwanted space

\thanks{Manuscript received April 19, 2005; revised August 26, 2015.}}


\markboth{Journal of \LaTeX\ Class Files,~Vol.~14, No.~8, August~2015}%
{Shell \MakeLowercase{\textit{et al.}}: Bare Demo of IEEEtran.cls for Computer Society Journals}



\IEEEtitleabstractindextext{%
\begin{abstract}
%Facial expression plays an important role in understanding human emotions. Most recently, deep learning based methods have shown promising for facial expression recognition. However, the performance of the current state-of-the-art facial expression recognition (FER) approaches usually degrades on unseen in-the-wild data because the models are prepared with training datasets generally collected under well-controlled settings and the number of labeled training samples is generally limited.
Facial expression plays an important role in understanding human emotions. Most recently, deep learning based methods have shown promising for facial expression recognition. However, the performance of the current state-of-the-art facial expression recognition (FER) approaches is directly related to the labeled data for training. To solve this issue, prior works employ the pretrain-and-finetune strategy, \textit{i.e.}, utilize a large amount of unlabeled data to pretrain the network and then finetune it by the labeled data. As the labeled data is in a small amount, the final network performance is still restricted. From a different perspective,  we propose to perform omni-supervised learning to directly exploit reliable samples in a large amount of unlabeled data for network training. Particularly, a new dataset is firstly constructed using a primitive model trained on a small number of labeled samples to select samples with high confidence scores from a face dataset,~\textit{i.e.}, MS-Celeb-1M \cite{guo2016ms}, based on feature-wise similarity. We experimentally verify that the new dataset created in such an omni-supervised manner can significantly improve the generalization ability of the learned FER model. However, as the number of training samples grows, computational cost and training time increase dramatically. To tackle this, we propose to apply a dataset distillation strategy to compress the created dataset into several informative class-wise images, significantly improving the training efficiency. We have conducted extensive experiments on widely used benchmarks, where consistent performance gains can be achieved under various settings using the proposed framework. More importantly, the distilled dataset has shown its capabilities of boosting the performance of FER with negligible additional computational costs. We hope this work will serve as a solid baseline and help ease future research in FER.

\end{abstract}

}

\maketitle

\IEEEdisplaynontitleabstractindextext

% \fi
\IEEEpeerreviewmaketitle

\IEEEraisesectionheading{\section{Introduction}\label{sec:introduction}}

\IEEEPARstart{A}{s} one of the most important emotion and intention sensing cues for human communications, facial activities have been extensively studied in the past decades. A robust facial expression recognition (FER) system should be able to recognize the basic expressions~\cite{lucey2010extended},~\textit{i.e.}, anger, disgust, fear, happy, sad, surprise, and/or compound expressions from expressive facial images. 
With the development of deep Convolutional Neural Networks (CNNs) \cite{7438833_tnnls2016,9463398_tnnls2021,fan2021unsupervised_tcyb2021}, various deep learning models  have been designed for facial activity analysis \cite{meng2017identity,Zhang_2018_CVPR,yang2018facial,8767026_tnnls2020,liu2021point_tcyb2021, zhang2021pro,zhang2021face_tip2021}.

As a data driven approach, CNNs require a huge amount of labeled data with high quality for training. However, collecting and annotating data with sufficient diversities is time-consuming and laborious. To solve this issue, researchers propose various ways to train deep models with limited data \cite{yan2016image_tmm2016,santander2021pitfalls,9425436_tnnls2021}. For example, Santander \textit{et al. }\cite{santander2021pitfalls} proposed to combine a group of small datasets into a large dataset to improve the performance. However, using such a dataset consisting of heterogeneous data for training requires a carefully designed training strategy, \textit{e.g.}, weights initialization, data augmentation etc to achieve a satisfactory result.

{Inspired by semi-supervised learning, 
%researchers in facial expression recognition resort their eyes to the huge amounts of facial images from Internet in order to solve the data shortage problem has been bothering them. However, how to take advantage of those huge-scale data as well as the labeled small size data is still under exploration. 
researchers have devoted efforts to studying how to take advantage of the huge amounts of unlabeled facial images from Internet for training FER systems, along with the existing datasets with limited labeled samples.
One plausible way of making use of those huge-scale unlabeled data, is to utilize them to pretrain a network~\cite{Li2020} in an unsupervised \cite{erhan2010does_icml2010} or self-supervised \cite{wang2021dense_cvpr2021} manner, and then use the pretrained network to finetuning based on the limited annotated data. However, as the dataset used in the finetuning step is of a small size, the knowledge learned by the network might be quite restricted. }

% Thus, we argue that an alternative, and (maybe) better way is to perform omni-supervised learning by exploiting the available labeled data~\textit{together} with large-scale unlabeled data to train the FER learner directly. As pointed out by~\cite{radosavovic2018data}, the performance of an omni-supervised learner, which utilizes both large-scale unlabeled data and the small size labeled data, can outperform its counterparts utilizing labeled data alone.}

{From a different perspective, in this work, we propose a simple yet effective baseline of omni-supervised facial expression recognition. In particular, under the omni-supervised learning setting, we select useful samples from a huge scale unlabeled data in an~\textit{automatic} manner. The selected samples is utilized to construct a large dataset, which is combined with existing small scale datasets for training the FER model, rather than for pretraining. Comparing to previous works following the pretraining-finetuning pipeline, our method has more advantages: it directly applies the large amount of samples selected from unlabeled data on network training, learning more abundant knowledge for the target task.}

% By our method, the statistical distribution gap between selected samples and the manually labeled data can be minimized. Therefore, the constructed dataset in a large size can provide strong auxiliary information, collaborates with the manually labeled data to train the FER learner. The effectiveness of the proposed approach has been demonstrated by the experimental results in terms of accuracy improvement for FER.}
%After discarding unaccountable samples, which might bring the noise to training, a huge scale labeled data (approximately 140K) with fine quality is constructed. Taking advantage of this constructed dataset as well as the~\textit{manually} labeled training set for training can greatly boost the recognition accuracy of the testing set, which has been proven by our experimental results.}

{Although the dataset constructed by the above mentioned method can significantly improve the performance of FER, which has been proved in our experiment, it still has issues. First, the size of the constructed dataset can be large (\textit{e.g.} hundreds of thousands of samples),  making the FER learner training process time-consuming. Second, since the constructed dataset is annotated automatically, there might exist some incorrectly labeled samples. Third, the number of selected samples in each class  might not be balanced, which might bring bias for training. To tackle the above-mentioned issues, we propose to utilize the dataset distillation strategy~\cite{wang2018dataset} to distill the key knowledge from the constructed large scale dataset. The distilled knowledge is compressed into a small set of images, \textit{e.g.}, in our case, one image represents one single category. Comparing to the original constructed dataset, the computational cost of the distilled images is decreased significantly. Although with a surprisingly small size, the distilled dataset is capable of boosting the recognition accuracy to a higher stage in both inner-dataset evaluation and cross-dataset evaluation comparing to the models trained using only the labeled datasets.}
% , which has been demonstrated in our experiment

% Despite the simplicity of our approach***

In general, we list our main contributions in this work as follows:
% \plnote{down below modified}
\begin{itemize}
% \item[-] An omni-supervised learning based approach is proposed and employed to construct an unlabeled dataset along with the labeled samples for training an FER model;

% \item[-] A dataset distillation strategy is presented and exploited to summarize the useful knowledge from the constructed unlabeled dataset to improve the generality, decrease the training cost, and balance the class number ratios;

% \item[-] Extensive experiments have been conducted on FER benchmarks, demonstrating the superiority of the proposed simple yet effective method.

\item[-] We propose an omni-supervised facial expression recognition baseline method. Specifically, we construct a large-scale dataset and exploit it under the omni-supervised pipeline. Unlike previous works utilizing an unlabeled large-scale dataset to pre-train the network and provide initialization for downstream fine-tuning, our method constructs a large-scale dataset from the huge amount of unlabeled data, and utilize the constructed dataset as well as the existing small scale labeled dataset to improve the FER performance.

\item[-] To improve the generality, decrease the training cost, and minimize training bias due to class imbalance, we propose to utilize the dataset distillation strategy to summarize the knowledge from the constructed dataset. Specifically, by the dataset distillation strategy, the constructed dataset, which has around 140K images, is compressed into a significantly small image set, in which one image for each class. The highly summarized knowledge in the distilled small set can be exploited together with the small size manually labeled data to improve the generality power of the FER learner. 

\item[-] We conduct extensive experiments on FER benchmarks, demonstrating the superiority of the simple yet effective baseline method.  

% We also conduct cross-dataset evaluations,~\textit{i.e.}, training a CNN on a source dataset but test on a different target dataset, which is quite challenging due to the well-know ``domain-shift" issues. Our performances in this cross-data setting demonstrate the superior generality of the proposed method. 
\end{itemize}

% ===================================================================================================
\section{Related Work}
This section will focus on facial expression recognition and semi-supervised learning, which are the most related topics with this article.

\subsection{Facial Expression Recognition}

With the development of modern convolutional neural networks, various deep learning methods have been applied to facial expression recognition. Recent deep learning based methods~\cite{Liu2014,mollahosseini2016going,meng2017identity,Zhang_2018_CVPR,yang2018facial,li2018occlusion} outperform their counterparts 
%without using deep learning 
 using traditional machine learning techniques~\cite{zhang1998comparison,zhang2005active,tian2002evaluation,eckhardt2009towards,yang2007boosting,hu2008multi,dahmane2011emotion,senechal2011combining,valstar2012meta,zafeiriou2010sparse,ying2010facial,liu2013improving,zhong2012learning,8101548, 8103056}.~\cite{Liu2014} proposes to unify Deep Belief Network and Boosting methods to perform feature learning, feature selection, and classifier construction in a joint way. \cite{mollahosseini2016going} constructs an Inception-wise network and achieves promising recognition rates on seven public facial expression databases. %For focusing on variations introduced by expressions,
 To obtain discriminative features, \cite{meng2017identity,Zhang_2018_CVPR,yang2018facial} propose to disentangle expression-sensitive knowledge and analyze the facial expressions based on the expression-sensitive knowledge. To suppress the variations introduced by identities, \cite{meng2017identity} proposes an identity-aware convotional neural network to differentiate the identity sensitive knowledge from expression sensitive knowledge. \cite{Zhang_2018_CVPR, 9197663} propose to utilize generative adversarial networks (GANs) to disentangle expression variations from pose variations.~\cite{yang2018facial} proposes to extract facial expressive information by a designed de-expression module. To recognize facial expressions with occlusions,~\cite{li2018occlusion} proposes to utilize attention mechanisms in CNNs and~\cite{Pan2019} utilizes a privileged learning mechanism to adjust the importance of feature learned from different facial regions. Interested readers can read~\cite{Huang2020} for a systematic review of deep facial expression recognition works. 

\subsection{Semi-supervised Learning}
A successful CNN based feature learning process is heavily dependent on large scale training data with high-quality labels. Unfortunately, manually labeling huge amounts of data in high quality is time-consuming and labor-expensive. More than that, in FER, due to the privacy concern, labeled data collection itself is also a question. To deal with this limitation, semi-supervised learning~\cite{zhu2005semi,tarvainen2017mean,laine2016temporal,DBLP:journals/corr/abs-1809-09925,radosavovic2018data,cai2018probabilistic} is utilized to improve the performance. Among those previous works,~\cite{cai2018probabilistic} follows a strategy that has been proved effective in previous works: introduce huge scale unlabeled data to pretrain the network, providing a good initialization for the downstream finetuning process. However, there are a few issues in this strategy~\cite{cai2018probabilistic}. Introduce huge-scale unlabeled data to pretrain the network occupies additional computation cost. The finetuning step still use a small scale dataset, which restricts the knowledge learned by the network. On the contrary, our method is based on omni-supervised learning, combining a huge-scale dataset, which is constructed automatically from a huge-scale unlabeled dataset, with a small size labeled dataset. By this way, the constructed dataset participates in the FER learner training process, rather than acting as an initializer. 

\section{Methodology}

% \subsection{Technical Details}
{In this section, we illustrate the details of the proposed method step by step. As shown in Fig.~\ref{fig:visualization-pipeline}, first, we utilize the manually labeled data in a small size to train a primitive FER learner (Sec.~\ref{subsec:primitive}). 
Second, this primitive FER learner is then utilized to provide guidance to select the most confident unlabeled data from huge-scale sets (Sec.~\ref{subsec:auxsample_collection}). Specifically, unlike previous works choosing the confident samples based on their category likelihood, this proposed baseline method is based on the similarity between high-level features extracted from the layer before the softmax output layer. Since this primitive FER learner embeds the knowledge from the manually labeled data in itself, the samples selected by it shares the similar statistical characteristics with the manually labeled ones. Consequently, the problem of ``domain shift" between the selected samples and the manually labeled samples can be relieved. Those huge scale selected samples are constructed into a new dataset, which will collaborate with the original small size labeled dataset to strengthen the FER learner directly. Further, to decrease the computation cost, save the training time, and increase the generality of the network, we propose to utilize the dataset distillation strategy~\cite{wang2018dataset} to compress the huge-scale constructed dataset into a very small set, in which one image for one class (Sec.~\ref{subsec:knowledge_dis})}.

\subsection{Task Definition}
Facial expression recognition aims to learn a mapping function to map the original input image $x$ to a latent representation. This latent representation should have the discriminative capability to distinguish samples from different classes. In this article, the mapping process is conducted by a convolutional neural network (CNN), which is denoted as $f(;\theta)$ and learned by labeled data $\{x_{i}^{l}, y_{i}^{l}\}, 1 \leq i \leq N$. The learning formulation is as follows:
\begin{equation}
   min \sum_{i=1}^{N} \sum_{j=1}^{m} \mathcal{L}(\textbf{w}_{j}^{T} \cdot f(x_{i}; \theta), y_i) 
%   + \alpha * \Omega((\theta))
   \label{eq:our-target}
\end{equation}
where $\mathcal{L}$(,) is a cross-entropy loss function, $\textbf{w}_{j}$ is the classifier parameter for $j$-th class, $\theta$ is the CNN parameter, m is the class number.
% $\Omega(.)$ is the regularization term.
% , and $\alpha$ is the trade-off parameter.

We target to improve the generality of the learned network $f(;\theta)$ by constructing a large scale dataset and apply it under the omni-supervised learning setting, which will be discussed in the next subsections.

% {In this work, we utilize knowledge distillation (Dis(.))  to compress the knowledge from the additional collected data labeled automatically,~\textit{i.e.}, 
% $\{x_{i}^{al}, y_{i}^{al} \}, 1 \leq i \leq n$. We fuse the distilled knowledge into the training process to boost the recognition power of the network. Since the knowledge is distilled and highly compressed, the introduced computation cost becomes negligible. Now the formulation becomes:}
% \begin{equation}
% %   min \sum_{i=1}^{N+n} L(W * f(x_{i}, y_i, Dis(x_i^{al}); \theta)) + \alpha * \Omega((\theta))
%   min \sum_{i}^{} L(W * f(x_{i}, y_i, Dis(x_i^{al}); \theta)) + \alpha * \Omega((\theta))
%   \label{eq:our-target}
% \end{equation}
% where $1 \leq i \leq N+n$.
% and $n \ll N$.

\subsection{Primitive Learner Training on Labeled Data}
\label{subsec:primitive}

\begin{figure*}[t]
\begin{center}
\includegraphics[width=1\linewidth]{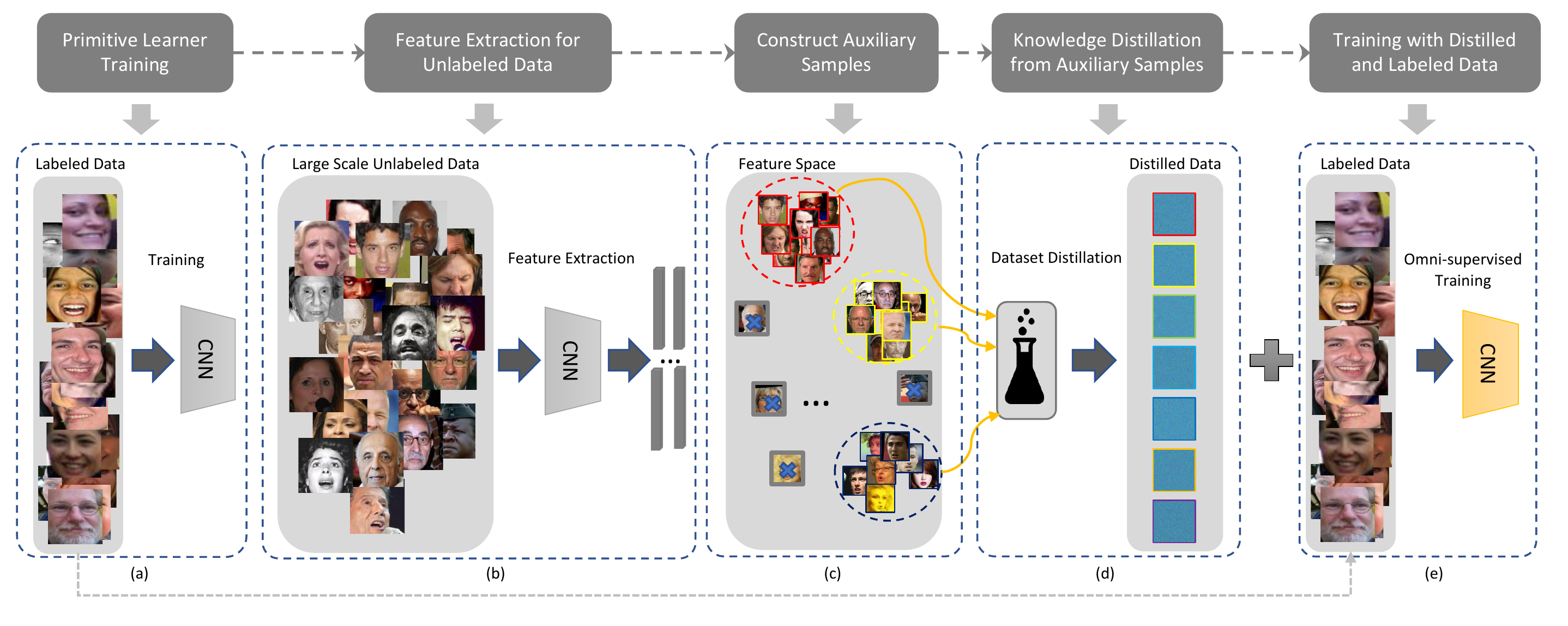}
\end{center}
\vspace{-4mm}

\caption{ An illustration for the whole pipeline. First, we utilize labeled data on hand to train a primitive classifier. Since the labeled dataset for FER is often with a small size, the computation cost in this step is not huge and acceptable. Second, this primitive classifier will be utilized to provide guidance to select the most related unlabeled data from large-scale sets.  Specifically, unlike previous works choosing the related samples based on their category likelihood, this proposed baseline method is based on high-level features with semantic meaning, which is extracted by the utilized primitive classifier. Knowledge related to FER will be injected into the training process, in which the useful information from unlabelled data and labeled data both contribute to the network training. Since the unlabelled data set probably consists of huge amounts of data, we utilize dataset distillation strategy~\cite{wang2018dataset} to compress the knowledge into a highly refined set,~\textit{i.e.}, one image for each class in our case. This highly distilled knowledge from huge-scale unlabeled data will be combined with the original labeled data together to train a stronger FER learner.}
\vspace{-4mm}
\label{fig:visualization-pipeline}
\end{figure*}

{As shown in Figure.~\ref{fig:visualization-pipeline}(a), in the first step, we utilize an existing small scale labeled dataset, denoted as $\{x_i^{l}, y_i^{l}\}$, to train a primitive learner $f(;\theta)$. To make the following discussion convenient, we name the utilized labeled dataset in this step as the anchor data. Given the anchor data, the primitive learner is trained by optimizing Eq. \ref{eq:our-target} via stochastic gradient descent \cite{johnson2013accelerating_nips2013}.}

% an input $x_{i}^{l}$, the trained primitive learner needs to predict the correct label $y_{i}^{l}$. }

% Our method is model-agnostic, so we can choose any architectures that have demonstrated their effectiveness in visual recognition problems~\cite{Parkhi15, he2016deep, hu2018squeeze,luo2019significance}. We do not need to modify the chosen architecture heavily or introduce any new loss terms to achieve better performances. The only light touch we need is to replace the final classifier layers and change the output number to the number of target facial expression classes. In this article, the output number is set to $7$ since there are seven basic facial expressions to detect,~\textit{i.e.}, anger, happy, disgust, fear, sad, surprise, and neutral. We need to emphasize that, in this step, the primitive learner is trained only via annotated facial expression data, which is in a limited size due to collection difficulty and annotation cost.} 
%  {To conduct a fair comparison, we follow the common practices as~\cite{meng2017identity, cai2018probabilistic},  choose VGG-Face~\cite{Parkhi15} and ResNet~\cite{he2016deep} to train our primitive learner}. 

% Because the manual annotation is highly labor-expensive, the number of the labeled samples is small, most of which are up to thousands. The network trained by a limited labeled data usually has a limited generality.} %The labeled data used for training the primitive learner is called Anchor Data (AD).
%########################################################################
%########################################################################

% , and SENet~\cite{hu2018squeeze} explicitly modeling inter-dependencies between feature channels

\subsection{Auxiliary Samples Selection from Unlabeled Data}
\label{subsec:auxsample_collection}
% This subsection illustrates how to improve the recognition power of the primitive FER learner by using the knowledge from huge-scale unlabelled data, which is denoted as $\{x_i^{ul}, y_i^{ul}\}$. 

This subsection illustrates how to select reliable samples  from a huge scale unlabeled dataset, which is denoted as $\{x_i^{ul}, y_i^{ul}\}$. To make the FER learner benefit from the selected unlabeled samples, the selected data should share similar statistical distributions with the labeled ones. Concretely, for each unlabelled data $x_i^{ul}$, 
we feed it to the primitive learner $f(;\theta)$ to generate its latent features $f(x_i^{ul}; \theta)$ and corresponding class confidence score by the following formulation:

% feed all the unlabeled data into $\phi_{primi}(w;.)$ to generate pseudo labels $y_*^{pl}$, as well as the latent representation $\textbf{f}_i$ in a high dimensional feature space. Then those original unlabeled data and their corresponding pseudo labels $\{x_i^{l}, y_i^{pl}\}$ can be fed into $\phi_{primi}(w;.)$ to update the model parameters. Concretely, to generate the pseudo label for an unlabelled data $x_i^{ul}$, a softmax classifier can be used to map the corresponding latent features $\textbf{f}_i$ into a probability score by the following formulation:
\begin{equation}
    P(y_i|f(x_i^{ul}; \theta)) = \frac{\mathrm{exp}(\textbf{w}_{i}^T \cdot f(x_i^{ul}; \theta))} {\sum_{j=1}^m\mathrm{exp}(\textbf{w}_{j}^T \cdot f(x_i^{ul}; \theta))},
\end{equation}
where $\textbf{w}_{j}$ is the weight vector for the $j$-th class, $m$ is the number of classes. 

% In our experiment, $m$ is set as $7$.

How to choose the appropriate samples sharing similar statistical distribution with the manually labeled data? One possible way is to utilize the probability score for the selection. Utilizing unlabeled data with pseudo labels generated by probability scores to update the learner has been explored in previous works on semi-supervised learning~\cite{zhu2005semi} and unsupervised learning~\cite{Luo_2019_CVPR}. However, in FER, the labeled data and unlabeled data are usually collected in different environments, making the conditional distributions between those databases different. As pointed out by~\cite{Li2020}, in such cases, the generated probability scores and corresponding pseudo labels, might be error-prone.
% and consequently, it might mislead the network learning process to an incorrect direction.
% if choose the unlabeled samples by their confidence score. 
% Under this circumstance, directly combining the unlabeled data with generated pseudo labels with the original labeled data might introduce unexpected noisy information for training the network, and consequently deteriorate the final performance

To deal this issue, we utilize the same strategy in~\cite{zhang2019category}: we generate pseudo labels for the unlabeled data based on the semantic feature similarity. As pointed out in~\cite{zhang2019category}, for samples $x_{*}$ in the same class, the corresponding learned latent features $f(x_{*}^{}; \theta)$ have a tendency to cluster together. The dimension of $f(x_{*}^{ul}; \theta)$ is usually much higher than that of probability score vector $P(y_{*}|x_{*}^{ul})$, making them more robust in learning processes. 

% Therefore, we follow the similar strategy in~\cite{zhang2019category}: select samples from the unlabeled data based on the high level feature similarity with the manually labeled samples.
% under the guidance provided by manually labeled data. 

Following the the strategy mentioned above, we feed each labeled sample~$x_{i}^{l}$ into the primitive network $f(; \theta)$ to calculate its latent feature representation~$f(x_i^{l}; \theta)$. Since we have the ground truth for those labeled data, we can calculate the centroid in the feature space for each class by the following formulation:
\begin{equation}
    center_k = \frac{\sum_{i}  \mathbbm{1}(y_i^{l}=k) \cdot  f(x_i^{l};\theta)} {\sum_{i} \mathbbm{1}(y_i^{l}=k)},
    \label{eq:center}
\end{equation}
where $\mathbbm{1}()$ is the indicator function, $k$ is the class index.  

As shown in Figure.~\ref{fig:visualization-pipeline}(b) and (c), for each unlabeled data sample $x_i^{ul}$, we feed it into the primitive FER learner to produce its feature representation $f(x_i^{ul};\theta)$, and then compute the cosine distance between $f(x_i^{ul};\theta)$ and each class centroid ~$center_k$: 
% In this work, we utilize the cosine distance with the following formulation:
% due to its??
\begin{equation}
   d(center_k, f(x_i^{ul};\theta)) = cos(center_k, f(x_i^{ul};\theta))
%   \frac{center_k \cdot f(x_i^{l};\theta)}{||center_k|| \ || f(x_i^{l};\theta)||}
%   d(center_k, f_{i}) = cosine(center_k, f_{i})= \frac{center_k \cdot f_{i}}{||center_k||||f_{i}||}
%   y_i^{ul} = \hat{k}, if  d_i^{\hat{k}} \leq d_i^{k} - m, \forall k \neq \hat{k}
\label{eq:distance}
\end{equation}

Then, we set the pseudo label for each unlabeled sample $x_{i}^{ul}$ based on the following criterion:
\begin{equation}
%   y_i^{ul} = \hat{k}, if \  d_i^{\hat{k}} \leq d_i^{k} - \delta, \forall k \neq \hat{k}
    y_i^{ul} = \hat{k}, If \  d(center_{\hat{k}}, f_{i}) \leq d(center_k, f_{i}) - \delta, \forall k \neq \hat{k}
   \label{eq:select}
\end{equation}
where $k$ is the class index for the facial expressions, $\delta$ is a margin threshold.

% Our hypothesis is that generated pseudo labels based on high dimensional semantic features are more robust and less error-prone. And therefore

The samples selected from the unlabeled dataset based on the above criterion would share the similar statistical distribution with the labeled samples. They collaborate with the existing small scale labeled data to construct a large scale dataset, which is utilized to refine the primitive FER learner $f(;\theta)$. To make the following discussion simple, we name those selected samples from an unlabeled database by our method as auxiliary samples. In our experiment, the number of auxiliary samples selected from the unlabeled MS1M-Celeb-1M (MS1M for short) is $137,769$. {We show examples of selected auxiliary samples in Figure.~\ref{fig:visualization-ms1m-selected-raf-based}, in which selected samples are sorted by their cosine distance values to the nearest class centroid.}

The efficacy of the auxiliary samples selected from the huge-scale unlabeled dataset are demonstrated in our experimental results, which will be discussed in Sec.~\ref{sec:inner-dataset} and~\ref{sec:cross-dataset}. 

% Comparing with the unlabeled data selected by their conditional likelihood score, generated pseudo labels based on high dimensional features are more robust and less error-prone. Those selected samples from the unlabeled database will be combined with those from the labeled database to help to update the model by ``learning with more trustable data". To make the following discussion simple, we name those selected samples from an unlabeled database as auxiliary samples. In our experiment, the number of auxiliary samples selected from the unlabeled MS1M-Celeb-1M (MS1M for short) is $137,769$. {We show examples of selected auxiliary samples in Figure.~\ref{fig:visualization-ms1m-selected-raf-based}, in which selected samples are sorted by their confidence values.}.
% The selected auxiliary samples collaborate with the manually labeled data by feeding to the primitive FER learner $f(;\theta)$ to refine the parameter $\theta$ and the classifier parameter $W$.

%\subsection{Auxiliary Sample Distillations}
\subsection{Knowledge Distillation from Auxiliary Samples}
\label{subsec:knowledge_dis}
\input{algorithm.tex}

Although the selected auxiliary samples can be used to boost the FER model performance, their large size brings significant computation cost \footnote{The size might keep increasing if we utilize more unlabeled data collected from the internet.}. Other than that, it is inevitable that some noisy samples still exist in those selected data. To minimize the computation cost and negative impact from noisy  auxiliary samples, we deploy dataset distillation~\cite{wang2018dataset}. Unlike model distillation, which distills knowledge from a set of separately trained learners into a compact one, dataset distillation conducts an orthogonal task: keep the learner fixed while distill the knowledge existing in a whole dataset and compress them into a sparse set of synthetic images. By an elegantly designed algorithm~\cite{wang2018dataset}, the distilled image set, although with a small size, has the capability to achieve the same or comparable performance.

Assume we have selected $N$ auxiliary samples, the target in this step is that we would like to encapsulate the knowledge in those $N$ samples into $n$ distilled samples, where $n \ll N$. Following~\cite{wang2018dataset}, we perform the dataset distillation process as detailed in Alg.~\ref{alg:dd}, by which we distill the knowledge from the huge size of auxiliary samples into a small set of synthetic images, denoted as $\tilde{\textbf{x}}$. We utilize the distilled $\tilde{\textbf{x}}$ as well as the existing small scale dataset,~\textit{i.e.}, $\{x_{i}^{l}\}$, to refine our FER learner.

{In summary, utilize the distilled auxiliary samples rather than the vanilla auxiliary samples  has a few advantages: 1) the distilled images are with a small size, introducing few computation cost; 2) comparing to the vanilla auxiliary collected samples, the distilled images are with high generality capability, which has been experimentally proved to benefit to the cross-dataset evaluation settings.}

% 3) the sample number of each class in distilled auxiliary data is balanced, without the potential of hurting the data balance when training.}
% ===================================================================================================
\section{Experimental Results}
\label{section:experiments_results}

In this section, we evaluate our method on benchmark datasets.  The details about evaluation settings, datasets,  implementation details, and performance comparisons are discussed in the following subsections.

\subsection{Evaluation Settings}
We employ two evaluation settings,~\textit{i.e.}, inner-database evaluation, and cross-database evaluation. In Inner-database evaluation, the training set and test set are from the same database. {For example, if we conduct an inner-evaluation on RAF-DB 2.0, we use RAF-DB 2.0 training set and constructed dataset to train an FER learner, and then use the trained learner for evaluating the RAF-DB 2.0 test set.} In Cross-dataset evaluation, the training set and test set are from different databases. 

% For example, one of our cross-database evaluation settings is as follows: the training set is from RAF-DB 2.0 and constructed dataset, while the testing set is from CK+.
% JAFFE, MMI, which are following~\cite{Li2020} for a fair comparison.

% and Static Facial Expressions in the Wild (SFEW) 2.0~\cite{dhall2011static}

\subsection{Databases}
% We employ the following two real-world datasets and three lab-controlled datasets to conduct a comprehensive evaluation for validating the efficacy of our method. We employ the following datasets to conduct comprehensive evaluations for validating the efficacy of our method. 
Experiments are conducted on five datasets, including Real-world Affective Face Database (RAF-DB) 2.0~\cite{li2018reliable}, FER-2013~\cite{goodfellow2015challenges},  Extended CohnKanade (CK+)~\cite{lucey2010extended}, Japanese Female Facial Expression (JAFFE)~\cite{lyons1998japanese}, and MMI~\cite{pantic2005web}. Auxiliary samples are selected from MS1M~\cite{guo2016ms}, which does not have any expression labels and therefore is considered as an unlabeled database. 
% RAF-DB 2.0~\cite{li2018reliable} contains 29,672 facial images which are collected from the Internet. It is a real-world database consisting of highly diverse samples. To achieve reliable labels for those samples, manually crowd-sourced annotation is conducted in~\cite{li2018reliable}. There are two emotion sets in this dataset, i.e. the basic expression label set, and the compound emotion label set. In this article, we focus on recognizing the basic expressions, and therefore we only utilize the basic expression label set, where there are 15,339 images divided into a training set (12,271 images) and a test set (3,068 images). 
    
    \textbf{RAF-DB 2.0}~\cite{li2018reliable} contains 29,672 facial images which are collected from the Internet. It is a real-world database consisting of highly diverse samples. To achieve reliable labels for those samples, manually crowd-sourced annotation is conducted in~\cite{li2018reliable}. There are two emotion sets in this dataset,~\textit{i.e.}, the basic expression label set, and the compound emotion label set. In this article, we focus on recognizing the basic expressions, and therefore we only utilize the basic expression label set, where there are 15,339 images divided into a training set (12,271 images) and a test set (3,068 images). To save the space, we use RAF-DB to denote RAF-DB 2.0 dataset in our discussion.
    
     \textbf{FER-2013}~\cite{goodfellow2015challenges} was constructed for the ICML 2013 Challenges in Representation Learning, which contains 28,709 training images, 3,589 validation images, and 3,589 test images with basic expression labels. After being collected from the Internet automatically by Google search engine, all images are aligned and resized to $48 \times 48$ pixels. 
     
     \textbf{CK+}~\cite{lucey2010extended} is a laboratory-controlled database that has been widely used in previous works for FER, where there are 593 video sequences collected from 123 subjects. In each sequence, there is a shift from a neutral expression in the first frame to the peak expression in the last frame. Among the 593 video sequences, there are 327 sequences labeled with basic expressions based on the Facial Action Coding System (FACS). Since CK+ does not provide official training/validation/test sets split, to make a fair comparison, we follow the setting in the previous work~\cite{Liu2014} to prepare the data. Specifically, first, we utilize the first frame as the neutral face of each labeled sequence and the last three peak frames with corresponding labels, resulting 1,308 images in total. Then the 1,308 images are divided into 10 groups for n-fold cross-validation experiments.
     
     \textbf{MMI}~\cite{pantic2005web} is constructed in a lab controlled environment, where there are 326 sequences in total, among which 213 sequences have basic expressions labels. Different from CK+, MMI is onset-apex-offset labeled, where the sequence in MMI begins with a neutral expression, reaches the peak expression in the middle, and then returns to a neutral expression. Pointed out by~\cite{Huang2020}, the subjects in MMI might perform the same expression in different ways, and occlusions,~\textit{e.g.}, glasses, mustache, exist in some of the subjects.
     
     \textbf{JAFFE}~\cite{lyons1998japanese} is a laboratory-controlled database containing 213 samples from 10 Japanese females, where each subject performs basic expressions 3 or 4 times. Due to the limited number of samples in this database, we utilize a leave-one-subject-out experimental setting.
    
    \textbf{MS-Celeb-1M(MS1M)}~\cite{guo2016ms} is a benchmark for the celebrity recognition. All images in MS1M are collected from the Internet. In version 1 of this dataset, there are around ten million images of one million celebrities captured in real scenarios. Due to its huge scale and variations, MS1M becomes one of the most challenging datasets for face recognition and face verification.  There are $85,742$ identities with    $5,822,653$ images in this pre-processed version, in which all the images are aligned by MTCNN~\cite{7553523} and resized to $112 \times 112$. There is no expression label in this database, and therefore we treat it as an unlabeled database in this work.
    % We utilize the data provided on https://github.com/ZhaoJ9014/face.evoLVe.PyTorch.

\begin{figure}[t]
\begin{center}
\includegraphics[width=1.0\linewidth]{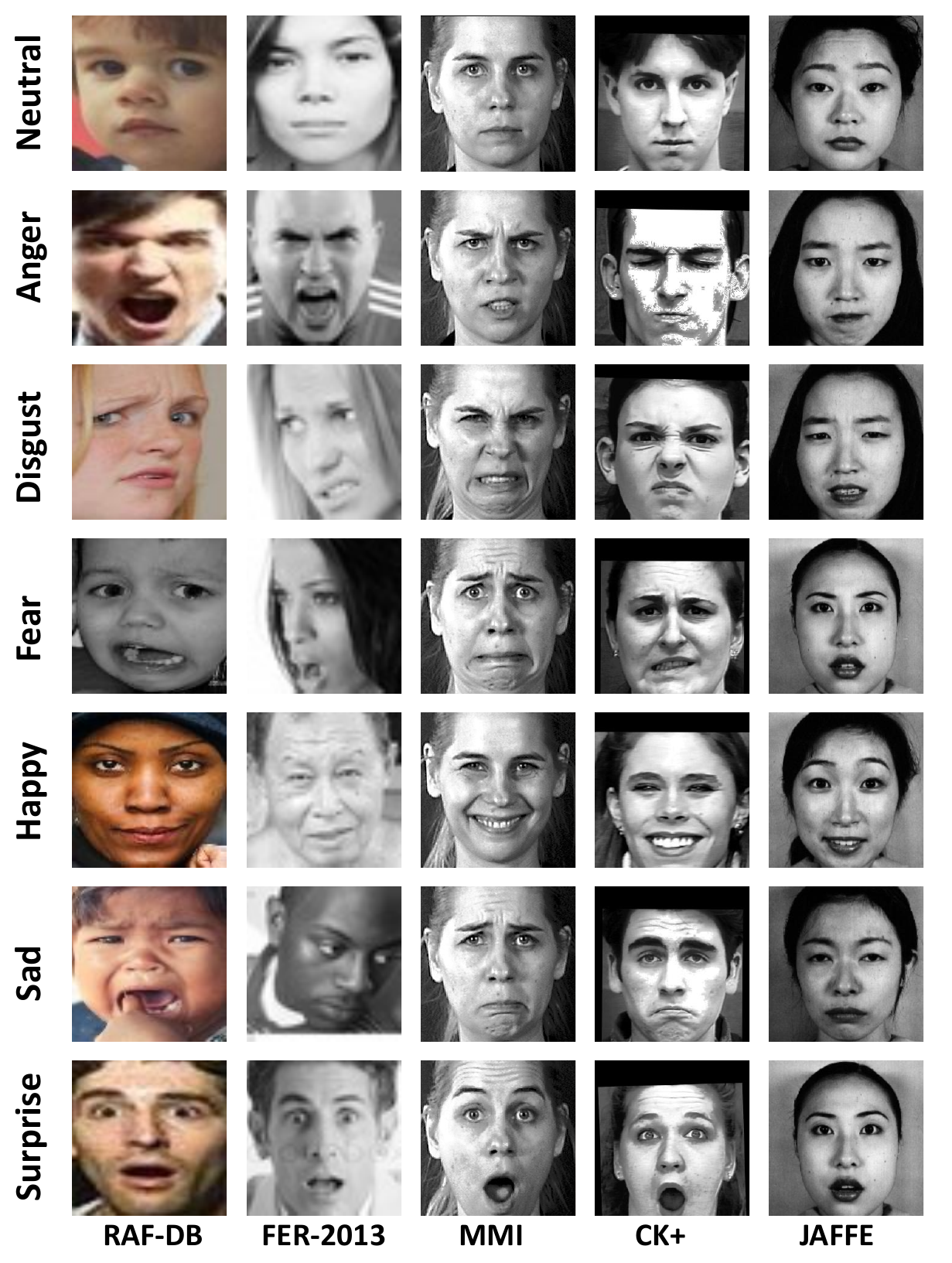}
\end{center}
\vspace{-4mm}

\caption{Visualizing examples from utilized databases. From top to bottom: neutral, anger, disgust, fear, happy, sad, and surprise. From left to right: RAF-DB 2.0, FER-2013, MMI, CK+, and JAFFE. Best viewed in color.}
\vspace{-4mm}
\label{fig:visualization-six-databases}
\end{figure}

In Figure.~\ref{fig:visualization-six-databases}, we show some example images from the databases utilized in this article. Those databases are collected in different scenarios, and therefore they have their own data bias. For example, 1) all faces in CK+ and JAFFE are frontal, while in RAF-DB, FER-2013, large face pose variations exist; 2) in FER-2013, CK+, and JAFFE, there is little color information to utilize; 3) the way to express ``anger" in RAF-DB, FER-2013 are quite different from that in CK+ and JAFFE.  We choose RAF-DB as the anchor database to select samples from MS1M, and use the selected samples to construct an auxiliary database for FER.
% in order to make a fair comparison with previous work~\cite{Li2020}.Based on the above observations, it is hard to directly utilize the data from the unlabeled database,~\textit{i.e.}, MS1M, to provide auxiliary information to train a network to achieve higher recognition accuracy on the labeled databases. 
%  Specifically, there are huge variations between different datasets, which are introduced by various factors,~\textit{e.g.}, illuminations, poses, subjects, scales.

%########################################################################
%########################################################################
%  RAF-DB 2.0 and FER-2013 are in-the-wild databases, while CK+ and JAFFE are datasets collected in the controlled environments.

% since the samples in RAF-DB2.0 are highly diverse. 
% We utilize a VGG-Face model to extract features for each sample in RAF-DB 2.0. 
% ===================================================================================================
\subsection{Implementation Details}
In our experiments, we utilize two different architectures,~\textit{i.e.},  VGG-Face~\cite{Parkhi15} and ResNet-34~\cite{he2016deep}, to test the efficiency and generality of our method. We choose those two architectures since they are the most frequently used networks in recent FER works~\cite{cai2018probabilistic,meng2017identity}.

All the images utilized in our experiments are aligned by MTCNN~\cite{7553523} and resized to $224 \times 224$. A data augmentation strategy of randomly horizontal flipping with 50\% probability is utilized. During training, we utilize stochastic gradient descent with 0.9 momentum to optimize the network. The initial learning rate is set to 0.001, which will be multiplied by 0.1 every ten epochs. Unless noted, the total epoch number is set to 25. We implement the experiments by PyTorch~\cite{NEURIPS2019_9015} and run all settings on a workstation with four NVIDIA GTX 2080Ti GPU cards.

% ===================================================================================================
% \subsection{Experimental Results}
\subsection{Performance Evaluation}
% {This section verifies the efficacy of our method in both discriminative ability and generality capability. First, to prove the effectiveness of our method, we conduct the inner-database evaluation, demonstrating the collected dataset and the corresponding distilled version can boost the recognition accuracy to a higher stage; second, we conduct a cross-dataset evaluation for proving the generality capability of our method. The performance is evaluated by the mean classification accuracy.}
This section verifies the efficacy of our method in both discriminative ability and generality capability. We conduct inner-database evaluation at first and cross-dataset evaluation at second. The performance is evaluated by the mean classification accuracy.
% and reported in  Sec.\ref{sec:inner-dataset} and \ref{sec:cross-dataset} respectively. 

% Under inner-dataset setting, we train a primitive learner by the training set of the given source database, and then utilize the primitive learner to guide the auxiliary sample selections in the unlabeled dataset. The knowledge from selected auxiliary samples from the unlabeled dataset will be injected into the learning process to improve the recognition accuracy, which is demonstrated in our experimental results,~\textit{i.e.}, Sec.\ref{sec:inner-dataset} and \ref{sec:cross-dataset}. To further improve the generality of the auxiliary samples while decreasing the introduced computational cost, we utilize dataset distillation strategy~\cite{wang2018dataset} to distill the knowledge from the selected auxiliary samples. The distilled auxiliary samples are highly expressive, which is one-image-one-class in our case. To make the discussion easier, we called the auxiliary samples before dataset distillation as vanilla auxiliary samples, the auxiliary samples after dataset distillation as distilled auxiliary samples.

\subsubsection{Inner-dataset Evaluations}
\label{sec:inner-dataset}

We conduct the inner-dataset evaluations on two in-the-wild databases,~\textit{i.e.}, RAF-DB, FER-2013 and one lab-controlled database,~\textit{i.e.}, CK+. We report comparisons with the prior works under the similar settings in Table.\ref{table:inner-RAF}, \ref{table:inner-FER-2013}, and \ref{table:inner-CK}. In the experiment, RAF-DB is utilized as the anchor dataset for training the primitive learner. Auxiliary samples are selected from MS1M. To save the space, the auxiliary samples after data distillation are denoted as DAS (Distilled Auxiliary Samples), while the auxiliary samples before distillation are denoted as VAS (Vanilla Auxiliary Samples).
% The selected auxiliary samples are shown in Figure.~\ref{fig:visualization-ms1m-selected-raf-based}.

In Table.~\ref{table:inner-RAF}, we report our performance on RAF-DB. Based on Table.~\ref{table:inner-RAF}, we can find that by utilizing the auxiliary samples selected from MS1M, the proposed method outperforms the previous works. Specifically, compared to the previous work utilizing new loss functions~\cite{li2017reliable}, our method does not need any new loss functions. Comparing to~\cite{cai2018probabilistic} introducing more human labels into training, our method still outperforms it by almost $2\%$ with few label cost. Since DAS is with a smaller size, its computation cost is much lower than VAS. The computation cost comparison is shown in Table.~\ref{table:visualization-computation-cost}. Comparing to the performance of VAS, the performance of DAS is still comparable, which is 86.55\% vs. 85.84\% on VGG-Face, 85.24\% vs. 85.84\% on ResNet-34, respectively. Considering the much less data size, the  distilled data achieves a good balance between computation cost and final performance. 
% More than that, it should be noted that the DAS is with a small size (7 in our experiment), and therefore the introduced additional computation cost is little, which is shown in Figure.~\ref{fig:visualization-computation-cost}. However, with little additional computational cost, the assistance brought by DAS is still comparable with that by VAS (86.55 vs. 85.84 on VGG-Face, 85.24 vs. 85.84 on ResNet-34).

Table.~\ref{table:inner-FER-2013} shows results comparison on FER-2013. On this challenging dataset, we find that the constructed dataset by our omni-supervised learning strategy and the distilled data both outperformed previous works under a similar setting. For example, comparing to the baseline with VGG-Face backbone, introducing the constructed dataset in an omni-supervised manner achieves a higher prediction accuracy; more than that, the distilled data boosts the performance to a higher stage by distilled key knowledge and keeping a sample balance from the constructed data.

% In this experimental setting, we test two anchor dataset cases, one is utilizing RAF-DB as the anchor dataset, and the other one is utilizing FER-2013 as the anchor dataset. From Table.~\ref{table:inner-FER-2013}, we can find that our method can achieve better or comparable results.

From Table.~\ref{table:inner-RAF} and~\ref{table:inner-FER-2013}, we can find that: 1) VGG-Face network has better results than ResNet-34. We believe this is because VGG-Face is pretrained on a face recognition dataset while ResNet-34 is initialized on ImageNet, which makes VGG-Face more suitable to face tasks. 2) distilled auxiliary samples achieve the best recognition accuracy in both settings,~\textit{i.e.} 86.55\% on RAF-DB and 73.27\% on FER-2013, which demonstrates that the distilled images contain key knowledge benefiting to the FER learner.

% when introducing additional knowledge from MS-Celeb-1M, our vanilla and distilled performance on VGG-Face are better (72.59 on VAS+VGG-Face+RAF-DB 2.0 as the anchor, 73.27 on DAS+VGG-Face+FER-2012 as the anchor) or comparable (72.12 on DAS + VGG-Face + RAF-DB 2.0 as the anchor, 72.08 on VAS + VGG-Face + FER-2013 as the anchor) compared with previous works. We also notice that under two settings, the performances are inferior to previous works by around 1\%, our conjectures is that this might comes from the large variations existed in this dataset, including poses, illuminations, and scales.   

% We also evaluate our method on CK+, which is the most widely used database collected in a lab-controlled environment. 
For CK+, we only conduct the experiment using VGG-Face and report the performance in Table.~\ref{table:inner-CK}. Since the data size of CK+ is much smaller than the size of vanilla auxiliary data, we do not conduct the experiment on CK+ with vanilla auxiliary samples, but only report the accuracy using distilled auxiliary samples. In Table.~\ref{table:inner-CK}, we can find that our method performs better than the baseline~\cite{cai2018probabilistic} and~\cite{mollahosseini2016going}, or achieves comparable accuracy to~\cite{cai2018probabilistic}. But unlike~\cite{cai2018probabilistic}, our method is with much less label cost.

% without introduction of any new loss terms, our proposed method utilizing the knowledge from seven distilled auxiliary sample boosts the accuracy in this lab-controlled dataset from 93.42\% (baseline) to 95.35\%, and is comparable with the previous state-of-the-art method using additional attribute information~\cite{cai2018probabilistic}.    

\subsubsection{Cross-dataset Evaluations}
\label{sec:cross-dataset}
The cross-dataset evaluations and comparisons with previous state-of-the-arts are shown in Table.~\ref{table:cross-dataset-ck}-\ref{table:cross-dataset-FER-2013}. Four datasets are used for test, which include: CK+, JAFFE, MMI, and FER-2013. We make a comparison with~\cite{Li2020}, which has the latest results on cross-dataset evaluation settings. Following the setting as~\cite{Li2020}, we use the combination of RAF-DB and distilled auxiliary samples selected from MS1M as training set. In this work, we do not compare to the prior methods utilizing specifically designed domain adaptation strategies due to two reasons: 1) our research focus is to demonstrate the efficacy of our omni-supervised FER method and the distilled data; 2) the domain adaptation strategies proposed in prior works for minimizing the domain gap between different datasets, such as \cite{luo2019significance,luo2020adversarial_nips2020,luo2021category_pami2021, tjio2021adversarial_wacv2021}, are complementary to our work and can collaborate together.

% We conduct cross-dataset evaluations on three lab-controlled databases,~\textit{i.e.}, CK+, JAFFE, and MMI, and one in-the-wild database,~\textit{i.e.}, FER-2013. The difference between our method and previous cross-database works is that our method only employes seven distilled auxiliary samples and combines them with the source dataset for training, without modifying the architecture, introducing new loss terms or using tons of extra training data. Since the size of the distilled auxiliary samples is small,~\textit{i.e.}, only $\textbf{seven}$, the introduced extra computational cost is negligible compared with the original training setting. In our experiment, VGG-Face is chosen as our backbone; RAF-DB 2.0 is utilized as the Anchor database, and distilled auxiliary samples (DAS) are selected from MS1M, RAF-DB, which are named as DAS-MS1M, DAS-RAF, respectively. 

Based on experimental results in Table.~\ref{table:cross-dataset-ck}-\ref{table:cross-dataset-FER-2013}, we can find that our method outperforms or achieves comparable results. We will give a more detailed analysis of the results for each dataset in the following paragraphs.
% from transferring to lab-controlled databases to transferring to in-the-wild databases, despite the fact that previous works utilize more datasets as the source domain or more complex network architectures.

%The performance comparison for cross-dataset evaluation on CK+ is shown in Table.~\ref{table:cross-dataset-ck}. 
As shown in the Table.~\ref{table:cross-dataset-ck}, our method achieves better performance than \cite{Li2020}. Unlike \cite{Li2020} using explicit domain adaptation strategy to minimize the domain gap between source dataset and target dataset, the better performance of our method is from the introduced distilled auxiliary samples with ignorable training cost.

% Note that CK+ is collected in a lab-controlled environment, and is quite similar to other lab-controlled datasets,~\textit{e.g.}, JAFFE, MMI, etc. Utilizing the lab-controlled datasets as source training datasets, as conducted in~\cite{mollahosseini2016going, hasani2017spatio}, was expected to minimize the impact of domain shift.~\cite{wen2017ensemble, wang2018unsupervised, Li2020} tried to utilize the datasets covering more variations to boost the generality of the trained network. However, compared with those previous works, our method improves the cross-dataset recognition accuracy on CK+ with only seven additional distilled auxiliary samples introduced in the training set without any network architecture modification or any additional loss/regularization term. Comparing with the latest work~\cite{Li2020} using the most similar setting with us, our method achieves a higher performance ($\sim$ 1.41\%). 

The performance on JAFFE dataset is reported in Table.~\ref{table:cross-dataset-jaffe}. As noted by \cite{Li2020}, there is a high bias in this extremely small dataset, which has only 213 images. Due to this bias, the cross-dataset evaluation on JAFFE is comparatively lower than that on CK+. However, our method still outperforms \cite{Li2020}. 

As shown in Table.~\ref{table:cross-dataset-mmi}, for MMI dataset, our method achieves comparable accuracy with \cite{Li2020} with few computation cost, outperforms the baseline by a large margin, which is from 60.82\% to 63.34\%.
% after introducing distilled auxiliary samples from unlabeled dataset. This performance improvement comes at a negligible computation cost without any network architecture modifications. ~\cite{zavarez2017cross} reported a higher accuracy using more extra data. In contrast, our method only employs source dataset for training.

We also conduct a cross-dataset evaluation on FER-2013. FER-2013 is constructed for challenges and has huge variations introduced by expressive ways, head poses, scales, and etc. Due to the heavy variations existed in FER-2013, the performance under the cross-dataset setting is inferior to that on CK+. In this challenging case, our method  still achieves a similar accuracy with~\cite{Li2020} with seven distilled auxiliary samples.

% From Table.~\ref{table:cross-dataset-FER-2013}, we can find that the knowledge extracted from 6 datasets~\cite{mollahosseini2016going} fails to provide the same assistance as the work using one in-the-wild database~\cite{Li2020}.

%########################################################################
\begin{figure}[t]
\begin{center}
\includegraphics[width=1\linewidth]{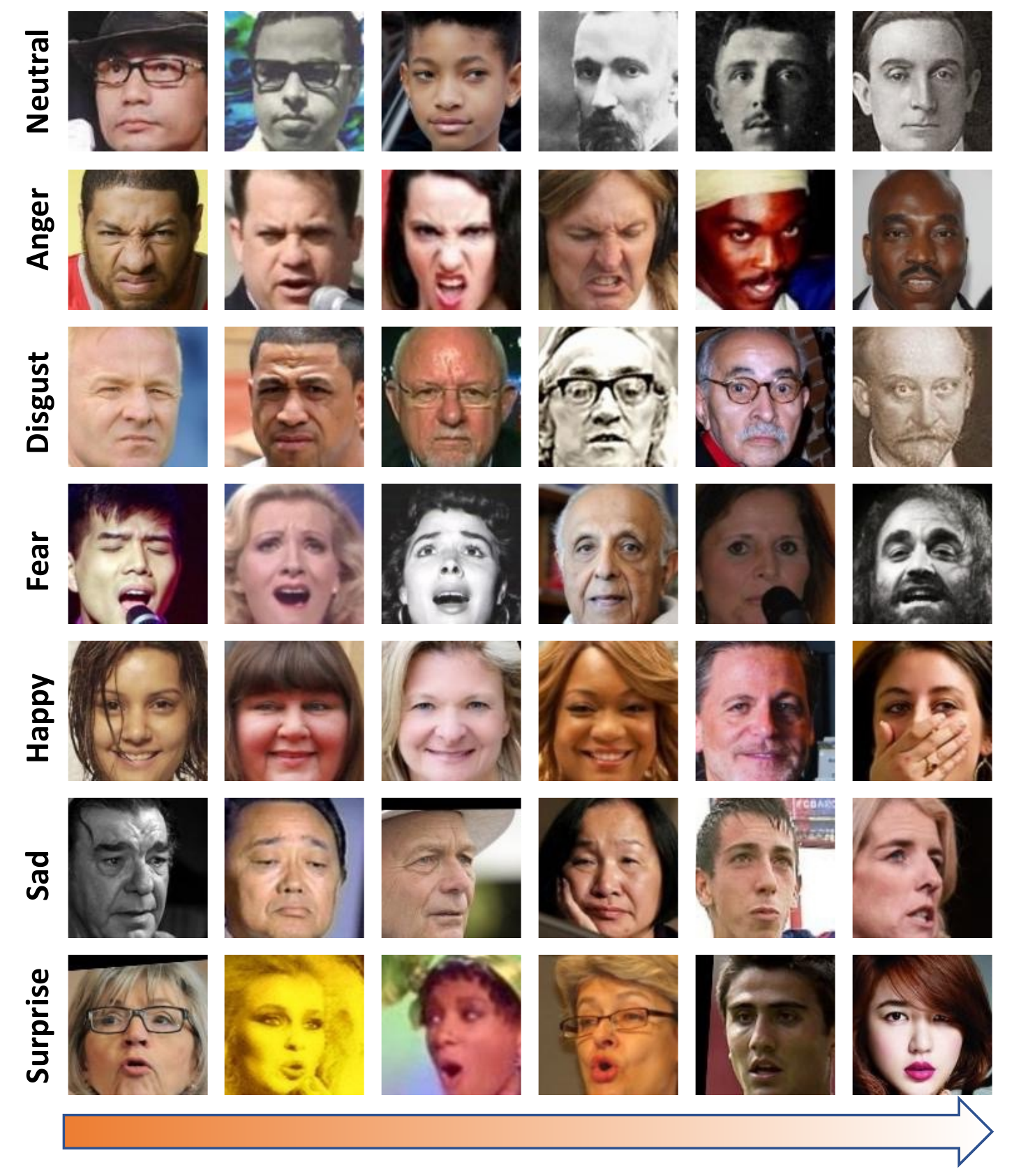}
\end{center}
\vspace{-4mm}
\caption{{Visualization of the selected auxiliary samples from MS1M database based on RAF-DB.} Each row corresponds to one expression. The confidences of images in the same row are decreasing. Best viewed in color.}
\vspace{-4mm}
\label{fig:visualization-ms1m-selected-raf-based}
\end{figure}
%########################################################################

%########################################################################
\begin{figure*}[t]
\begin{center}
\includegraphics[width=1.0\linewidth]{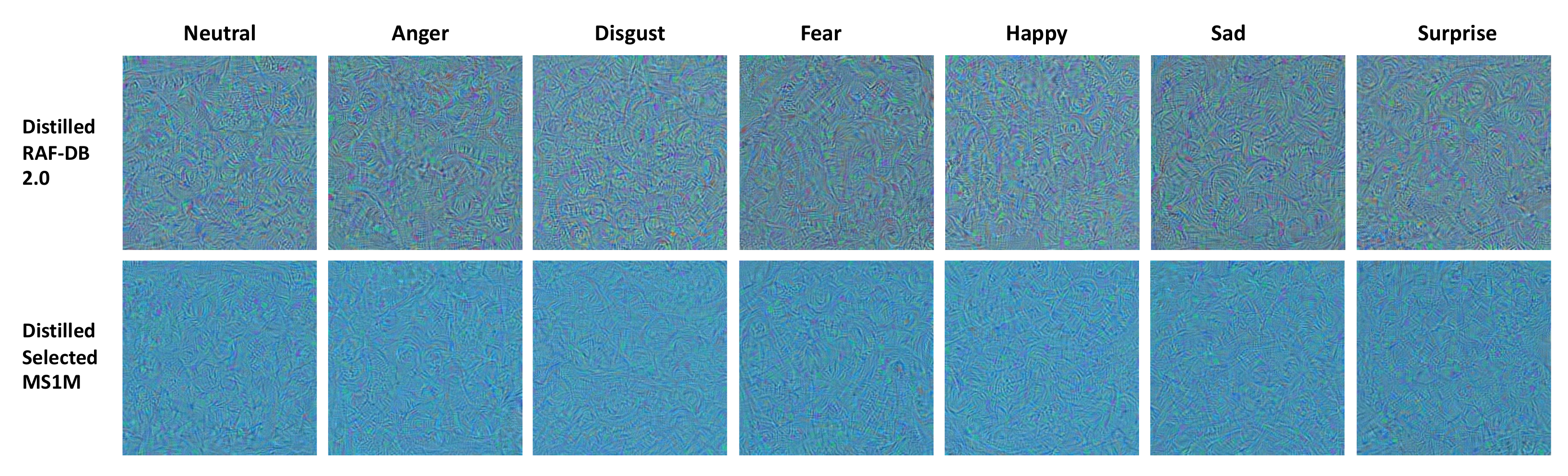}
\end{center}
\vspace{-4mm}
\caption{{Visualization of the distilled images from RAF-DB training set, and distilled images from selected samples in MS1M database.} Best viewed in color.}
% \vspace{-4mm}
\label{fig:visualization-distilled-samples}
\end{figure*}
\begin{table}[t]\setlength{\tabcolsep}{25pt}
\caption{Inner-dataset comparison of our method with previous works on RAF-DB. VAS stands for vanilla auxiliary samples, DAS stands for distilled auxiliary samples.} 
%The performance of previous works are cited from \cite{li2019pooling}.
\vspace{-2mm}
\label{table:inner-RAF}
\centering
\begin{tabular}{lc}
\toprule
% \multirow{2}{*}{Methods}                & \multicolumn{4}{c}{MARS}                      \\ 
Method                    & Accuracy (\%)      \\ \midrule %\midrule
FSN\cite{zhao2018feature}      & 81.10              \\
MRE-CNN\cite{fan2018multi}        & 82.63             \\
baseDCNN\cite{li2017reliable}        & 82.66                \\
DLP-CNN\cite{li2017reliable}         & 82.84               \\
Center Loss\cite{li2017reliable} & 82.86      \\
PAT-VGG-F-(gender,race)\cite{cai2018probabilistic}  &83.83 \\
PAT-ResNet-(gender,race)\cite{cai2018probabilistic}  &84.19 \\
% APM-VGG.\cite{li2019pooling}    & 85.17     (not cite)         \\  \midrule
VGG-F (baseline) & 85.15\\
ResNet-34 (baseline) & 84.63 \\
% ResNeXt-101 (baseline) & 85.85 \\
% SENet (baseline) & 0 \\ 
\midrule
% Chen et al.\cite{}        & 86.3                \\ \midrule
    % \midrule
    % \multicolumn{1}{c}{RAF-DB 2.0 as Anchor, select ASs from MS1M} \\
    % \midrule
\textbf{VAS + VGG-F }                       & \textbf{85.84}   \\  
\textbf{VAS + ResNet-34 }                       & \textbf{85.84}   \\ 
% Vanilla auxiliary samples + SENet (Ours)                       & 0   \\ 
% \midrule  
\textbf{DAS + VGG-F }                       & \textbf{86.55}   \\  
\textbf{DAS + ResNet-34 }                       & \textbf{85.24}  \\ 
% DAS + ResNeXt-101 (Ours) & 85.97  ??\\

\bottomrule
\end{tabular}
% \vspace{-4mm}
\end{table}
%########################################################################

%########################################################################
\begin{table}[t]\setlength{\tabcolsep}{25pt}
\caption{Inner-dataset comparison of our method with previous works on FER-2013. The performance of previous works are cited from \cite{cai2019improving}. }

\vspace{-2mm}
\label{table:inner-FER-2013}
\centering
\begin{tabular}{lc}
\toprule
% \multirow{2}{*}{Methods}                & \multicolumn{4}{c}{MARS}                      \\ 
Method                    & Accuracy (\%)       \\ 
\midrule
ECNN\cite{wen2017ensemble}        & 69.96             \\
DLSVM.\cite{tang2013deep}        & 71.2                \\
Ron et al..\cite{breuer2017deep}         & 72.1               \\
PAT-VGG-F-(gender,race).\cite{cai2018probabilistic}  &72.16 \\
PAT-ResNet-(gender,race).\cite{cai2018probabilistic}  &72.00 \\ 
% \midrule
VGG-F (baseline) & 71.79\\
ResNet-34 (baseline) & 70.89 \\
 
\midrule

% \multicolumn{1}{l}{\emph{RAF-DB as anchor database:}} \\
% \textbf{VAS + VGG-F (Ours)}                       & \textbf{72.59}   \\  
% \textbf{VAS + ResNet-34 (Ours)}                       & \textbf{70.31}   \\ 
% \textbf{DAS + VGG-F (Ours)}                       & \textbf{72.12}  \\  
% \textbf{DAS + ResNet-34 (Ours)}                       & \textbf{71.23}  \\ 

%     \midrule
% \multicolumn{1}{l}{\emph{RAF-DB as anchor database:}} \\
% \multicolumn{1}{l}{\emph{RAF-DB as anchor database:}} \\
\textbf{VAS + VGG-F }                       & \textbf{72.08}  \\  
\textbf{VAS + ResNet-34 }                       & \textbf{70.98}   \\ 
\textbf{DAS + VGG-F}                       & \textbf{73.27}   \\  
\textbf{DAS + ResNet-34}                       & \textbf{71.20}  \\ 

\bottomrule
\end{tabular}
% \vspace{-4mm}
\end{table}

%########################################################################

%########################################################################
\begin{table}[t]\setlength{\tabcolsep}{19pt}
\caption{Inner-dataset comparison of our method with previous works on CK+. The performance of previous works are cited from \cite{cai2018probabilistic}. }

\vspace{-2mm}
\label{table:inner-CK}
\centering
\begin{tabular}{lc}
\toprule
% \multirow{2}{*}{Methods}                & \multicolumn{4}{c}{MARS}                      \\ 
Method                    & Accuracy (\%)       \\ \midrule
Inception\cite{mollahosseini2016going}      & 93.20              \\

PAT-VGG-F-(gender,race)\cite{cai2018probabilistic}  &95.58 \\

VGG-F (baseline)~\cite{cai2018probabilistic} & 93.42\\

\midrule  
\textbf{DAS + VGG-F}                       & \textbf{95.35}   \\

\bottomrule
\end{tabular}
\vspace{-4mm}
\end{table}
%########################################################################

%########################################################################
\begin{table}[t]\setlength{\tabcolsep}{8pt}

\caption{Cross-dataset comparison on CK+ dataset. DA denotes domain adaptation. }
% ``6 Datasets" in~\cite{mollahosseini2016going} means MultiPIE, MMI, DISFA, FERA, SFEW, and FER-2013
\vspace{-2mm}

\label{table:cross-dataset-ck}
\centering
\begin{tabular}{lllc}
\toprule
% \multirow{2}{*}{Methods}                & \multicolumn{4}{c}{MARS}                      \\ 
Method      & Source           &Target      & Accuracy  (\%)      \\ \midrule 
% Mollahosseini\cite{mollahosseini2016going}    &6 Datasets    &CK+  & 64.2            \\ 

% Hasani et\cite{hasani2017spatio}    & MMI+JAFFE    &CK+  & 73.91             \\ 
% Wen\cite{wen2017ensemble}    & FER-2013    &CK+  & 76.05             \\ 
% Wang\cite{wang2018unsupervised}    &FER-2013    &CK+  & 76.58             \\ 
VGG-F+DA\cite{Li2020}    &RAF-DB    &CK+  & 78.00             \\ 
\textbf{VGG-F}    & \textbf{RAF-DB+DAS}    &\textbf{CK+} &\textbf{79.33}                \\   
% Li\cite{Li2020}    &RAF-DB2.0  &CK+  & 54.26             \\
% DAS-RAF + VGG-F     &RAF-DB2.0  &CK+  &79.33                \\   
% % Li\cite{Li2020}    &RAF-DB2.0  &CK+   & 55.38             \\
% DAS-MS1M+DAS-RAF + VGG-F     &RAF-DB2.0   &CK+  &79.41 \\

% \\  \midrule  \midrule

\bottomrule
\end{tabular}

% \vspace{-4.9mm}
\end{table}
% 6 Datasets: MultiPIE, MMI, DISFA, FERA, SFEW, and FER-2013

%########################################################################

%########################################################################
\begin{table}[t]\setlength{\tabcolsep}{8pt}

\caption{Cross-dataset comparison on JAFFE dataset. }
% ``6 Datasets" in~\cite{zavarez2017cross} means CK+, MMI, RaFD, KDEF, BU3DFE and ARFace.

\vspace{-2mm}

\label{table:cross-dataset-jaffe}
\centering
\begin{tabular}{lllc}
\toprule
% \multirow{2}{*}{Methods}                & \multicolumn{4}{c}{MARS}                      \\ 
Method      & Source           &Target      & Accuracy (\%)       \\ \midrule 

% Zavarez\cite{zavarez2017cross}    & 6 Datasets    &JAFFE  & 44.32             \\ 

% Wen\cite{wen2017ensemble}    & FER-2013    &JAFFE  & 50.70             \\ 
% Ali\cite{ali2016boosted}    &RaFD    &JAFFE  & 48.67             \\ 
VGG-F+DA\cite{Li2020}    &RAF-DB    &JAFFE  & 54.26             \\ %\midrule
\textbf{VGG-F}     & \textbf{RAF-DB+DAS}    &\textbf{JAFFE} & \textbf{54.93}                \\   
% Li\cite{Li2020}    &RAF-DB2.0  &CK+  & 54.26             \\
% DAS-RAF+VGG-F     &RAF-DB2.0  &JAFFE  &57.28                \\   
% % Li\cite{Li2020}    &RAF-DB2.0  &CK+   & 55.38             \\
% DAS-MS1M+DAS-RAF+VGG-F     &RAF-DB2.0   &JAFFE  &54.93 \\

% \\  \midrule  \midrule

\bottomrule
\end{tabular}

% \vspace{-4mm}
\end{table}

%########################################################################

%########################################################################
\begin{table}[t]\setlength{\tabcolsep}{8pt}

\caption{Cross-dataset comparison  on MMI dataset. }
% The first ``6 Datasets" in~\cite{zavarez2017cross} means CK+, MMI, RaFD, KDEF, BU3DFE and ARFac, the second ``6 Datasets" in~\cite{mollahosseini2016going} means MultiPIE, CK+, DISFA, FERA, SFEW, and FER-2013.
\vspace{-2mm}

\label{table:cross-dataset-mmi}
\centering
\begin{tabular}{lllc}
\toprule
% \multirow{2}{*}{Methods}                & \multicolumn{4}{c}{MARS}                      \\ 
Method      & Source           &Target      & Accuracy (\%)       \\ \midrule 

% Zavarez\cite{zavarez2017cross}    & 6 Datasets    &MMI  & 67.03             \\ 

% Mollahosseini\cite{mollahosseini2016going}    &6 Datasets   &MMI  & 55.6 \\

% Wang\cite{wang2018unsupervised}    &FER-2013    &MMI  & 61.86             \\ 
VGG-F+DA\cite{Li2020}    &RAF-DB    &MMI  & 64.13             \\ %\midrule
VGG-F (baseline)    &RAF-DB    &MMI  & 60.82             \\ %\midrule

% DAS-RAF+VGG-F     &RAF-DB  &MMI  &63.7                \\   

\textbf{VGG-F}     & \textbf{RAF-DB+DAS}   & \textbf{MMI}  & \textbf{63.34} \\
% \\  \midrule  \midrule

\bottomrule
\end{tabular}

\vspace{-4mm}
\end{table}

%########################################################################

%########################################################################
\begin{table}[t]\setlength{\tabcolsep}{8pt}
% \caption{Comparison of our method with previous works on cross-dataset evaluations on CK+ dataset. The performance of previous works are cited from~\cite{Li2020}. RAF-DB 2.0 is utilized as the Anchor database. distilled ASs selected from MS1M, RAF-DB, and FER-2013 respectively, 

% % which are named as AS_MS1M, AS_RAF, and AS_FER, respectively. VGG-Face as the backbone.
% }
\caption{Cross-dataset comparison on FER-2013 dataset. }
% ``6 Datasets" means MultiPIE, CK+, DISFA, FERA, SFEW, and FER-2013
\vspace{-2mm}

\label{table:cross-dataset-FER-2013}
\centering
\begin{tabular}{lllc}
\toprule
% \multirow{2}{*}{Methods}                & \multicolumn{4}{c}{MARS}                      \\ 
Method      & Source           &Target      & Accuracy (\%)        \\ \midrule 

% Mollahosseini\cite{mollahosseini2016going}    & 6 Datasets    &FER-2013  & 34.0             \\ 
VGG-F+DA\cite{Li2020}    &RAF-DB    &FER-2013  & 55.38             \\ %\midrule
\textbf{VGG-F}     & \textbf{RAF-DB+DAS}   & \textbf{FER-2013} &\textbf{55.17}                \\   
% % Li\cite{Li2020}    &RAF-DB2.0  &CK+  & 54.26             \\
% DAS-RAF+VGG-F     &RAF-DB+DAS  &FER-2013  &55.06                \\   
% % Li\cite{Li2020}    &RAF-DB2.0  &CK+   & 55.38             \\
% DAS-MS1M+DAS-RAF+VGG-F     &RAF-DB2.0   &FER-2013  &54.72 \\

% \\  \midrule  \midrule

\bottomrule
\end{tabular}

\vspace{-4mm}
\end{table}

%########################################################################

%########################################################################
\begin{table}[t]\setlength{\tabcolsep}{12pt}

\caption{Computation cost of without introducing auxiliary samples, with vanilla auxiliary samples, with distilled auxiliary samples. }
% ``6 Datasets" in~\cite{mollahosseini2016going} means MultiPIE, MMI, DISFA, FERA, SFEW, and FER-2013
\vspace{-2mm}

\label{table:visualization-computation-cost}
\centering
\begin{tabular}{lccc}
\toprule
% \multirow{2}{*}{Methods}                & \multicolumn{4}{c}{MARS}                      \\ 
Method      & w/o AS           &w/ VAS      & w/ DAS       \\ \midrule 
Seconds per epoch    &155    &1625   & 156             \\ 

\bottomrule
\end{tabular}
% \vspace{-4.9mm}
\end{table}

\subsubsection{Visualization of Selected Samples}
We visualize selected auxiliary examples from MS1M in Figure.~\ref{fig:visualization-ms1m-selected-raf-based}, and the distilled auxiliary samples in Figure.~\ref{fig:visualization-distilled-samples}. In total, our method selects $117,369$ auxiliary samples from MS1M.

Each row of Figure.~\ref{fig:visualization-ms1m-selected-raf-based} corresponds to one facial expressions. The images in each row are sorted by the distance between their feature and the nearest class centroid. Although there are large variations in MS1M, and a large domain gap between MS1M and the anchor dataset (RAF-DB), we can observe that the auxiliary samples selected are still highly related to activated facial expression. From the images shown in Figure.~\ref{fig:visualization-ms1m-selected-raf-based},  despite there are heavy illumination changes (the first column, second row), large poses (the first column and third column, sixth row), occlusions (fourth column, first row), our method still assigns correct expression labels for those samples. The high quality of those selected samples can explain why our method achieves a better performance after utilizing those auxiliary data.
% The selected auxiliary samples from the unlabeled database are consequently playing a significant role to update the primitive learner, which has been demonstrated in Sec.~\ref{sec:inner-dataset} and~\ref{sec:cross-dataset}. 
\begin{figure}[H]
\begin{center}
\includegraphics[width=0.95\linewidth]{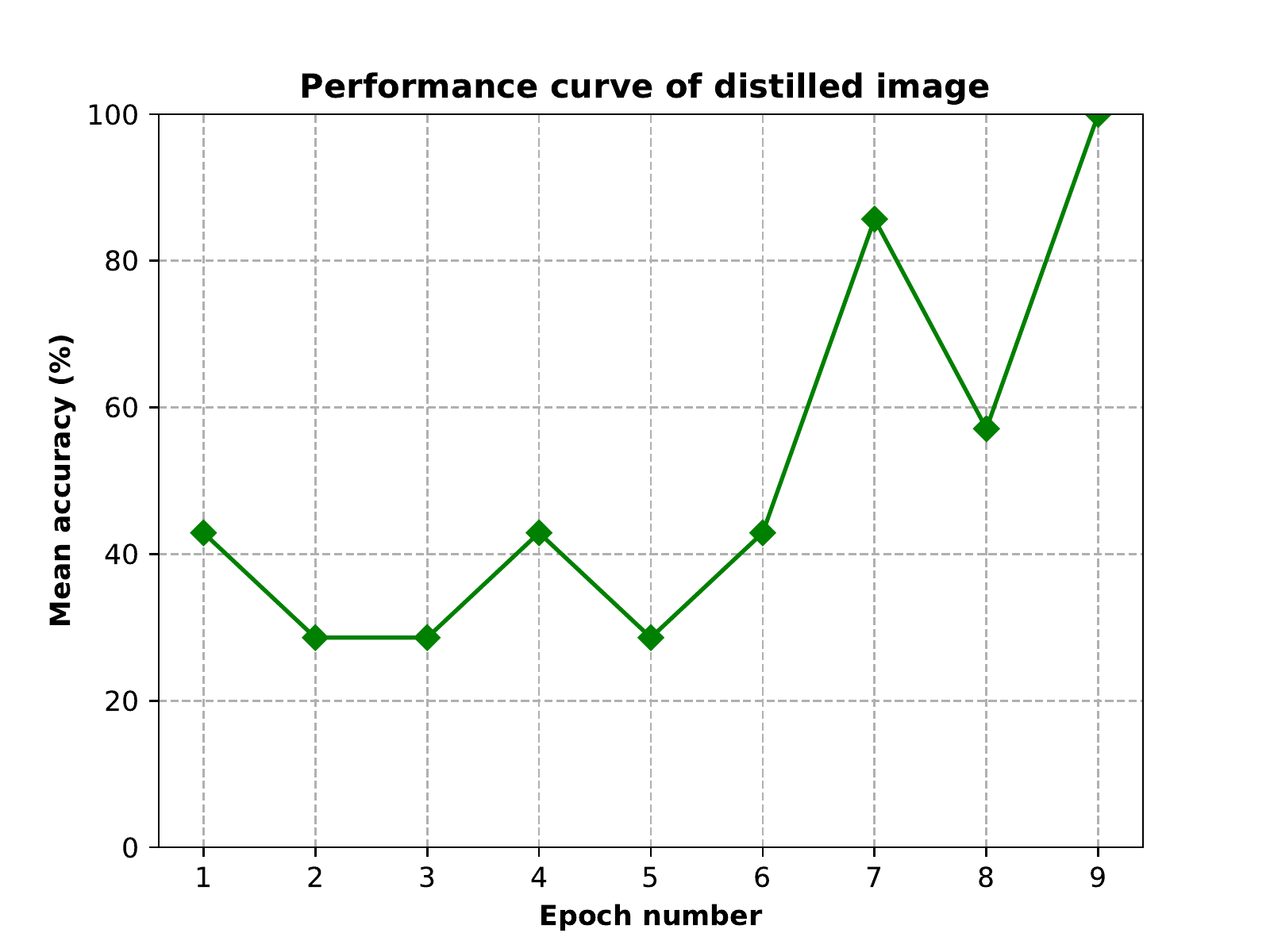}
\end{center}
% \vspace{-4mm}
\caption{Prediction curve of test set in distilled auxiliary samples. RAF-DB as the anchor dataset.}
\label{fig:visualization-underlying-patterns}
\end{figure}

\subsection{Discussion}
\textbf{Do the distilled auxiliary samples really contain underlying patterns?} Based on the visualizations in Figure.~\ref{fig:visualization-distilled-samples}, it is difficult to find any semantic meaning in them. All of the distilled images are full of specific texture patterns.  This observation arouses us the question that whether those distilled images can provide useful knowledge for us. We conduct a simple experiment to test whether or not there are any underlying class related patterns in those distilled images. We save the intermediate distilled images in every ten epochs, resulting in 42 distilled images (the distillation process takes six epochs, each epoch saves 7 images). We split them into a training set and testing set by a ratio of 5:1, and then feed the split training set and testing set to train a CNN. Our intuition is this: if those distilled auxiliary samples do not have any regular patterns corresponding to classes, then the CNN can not learn anything from them. What we observe is shown in Figure.~\ref{fig:visualization-underlying-patterns}. From this figure, we can find that the CNN trained on the distilled auxiliary samples converges very fast. The CNN can easily predict the images in the test set correctly, which indicates the existence of underlying patterns corresponding to different classes in each distilled sample.

\section{Conclusion}
In this article, we propose a simple yet effective omni-supervised baseline for facial expression recognition. Unlike previous works using unlabeled data for pretraining, we exploit and distill the useful knowledge from a large-scale unlabeled data set constructed by our method, and enhance the FER performance. We have demonstrated that the distilled knowledge from the constructed large-scale unlabeled data has high generality by achieving significant advancement in inner-dataset and cross-dataset evaluations. In the future, we will explore how to make our proposed method to collaborate the latest FER works such as PASM \cite{liu2021point_tcyb2021} , improving the network performance from different aspects.

% or


% \appendices

% % you can choose not to have a title for an appendix
% % if you want by leaving the argument blank
% \section{}

% % use section* for acknowledgment
% \ifCLASSOPTIONcompsoc
%   % The Computer Society usually uses the plural form
%   \section*{Acknowledgments}
% \else
%   % regular IEEE prefers the singular form
% \fi


\ifCLASSOPTIONcaptionsoff
  \newpage
\fi

\bibliographystyle{IEEEtran}
\bibliography{egbib}

\end{document}

%% file: algorithm.tex
% \begin{algorithm}[t]
% \caption{Dataset Distillation}
% \label{alg:dd}

% % \begin{algorithmic}[1]
% \textbf{Input}: the selected auxiliary samples with size $N$; \\
%      $p(\theta)$ : distribution of initial network weights; \ $\etal_{0}$: initialization for $\etal$ ;\\
%      $\alpha$ : step size; \ $n$: batch size; \ $T$: the iteration number;

% \State  \textbf{Given}:  pruning rate $P_{i}$
% \State  \textbf{Initialize}: model parameter $\mathbf{W}$
% % \For{$epoch=1$; $epoch \leq epoch_{max}$; $epoch++$}
% % 	\State Update the model parameter $\mathbf{W}$ based on $\mathbf{X}$
% % 	\State Find $\mathcal{A}^{*}$ that satisfy Eq.~\ref{eq:min-attribute}

% % 	\For{$i=1$; $i \leq L $; $i++$}
% % 		\State Using $\mathcal{A}^{*}$  to prune $N_{i+1}P_i$ filters
% % 	\EndFor
% % \EndFor
% % \State Obtain the compact model $\mathbf{W} ^{*}$ from $\mathbf{W}$
% % \OUTPUT The compact model and its parameters $\mathbf{W} ^{*}$
% % \end{algorithmic} 
% \end{algorithm}

\begin{algorithm}[t]
\caption{Dataset Distillation Process}
\label{alg:dd}

\begin{algorithmic}[1]
\INPUT the selected auxiliary samples with size $N$; \ $n$ : the distilled sample number; \ $p(\theta)$ : distribution of initial network weights; \ $\eta_{0}$: initialization for $\eta$ ; \ $\alpha$ : step size; \ $M$: batch size; \ $T$: the iteration number;
\State  \textbf{Initialize}: $\tilde{\textbf{x}}=\{ \tilde{x}_{i}\}_{i=1}^{n}, \eta=\eta_{0}$
\For{$t=1$; $t \leq T$; $t++$}
	\State fetch the current batch data $\textbf{x}_t = \{ x_{t,j}\}_{j=1}^{M} $
	\State sample a batch of initial weights $\theta_{0}^{j}$ based on distribution $p(\theta_{0})$

	\For{each sampled $\theta_{0}^{j}$}
		\State $\theta_{1}^{j}=\theta_{0}^{j}-\eta \triangledown_{\theta_{0}^{j}} l(\tilde{\textbf{x}}, \theta_{0}^{j})$
		\State $L^{j} = l(\tilde{\textbf{x}}_{t}, \theta_{1}^{j})$
	\EndFor
	\State Update: $\tilde{\textbf{x}}=\tilde{\textbf{x}}-\alpha \triangledown_{\tilde{\textbf{x}}} \sum_{j}L^{j}$, \ $\eta = \eta - \alpha \triangledown_{\eta}  \sum_{j}L^{j}$
\EndFor
\OUTPUT distilled data $\tilde{\textbf{x}}$
\end{algorithmic} 
\end{algorithm}

%% file: egbib.bbl
\begin{thebibliography}{10}
\providecommand{\url}[1]{#1}
\csname url@samestyle\endcsname
\providecommand{\newblock}{\relax}
\providecommand{\bibinfo}[2]{#2}
\providecommand{\BIBentrySTDinterwordspacing}{\spaceskip=0pt\relax}
\providecommand{\BIBentryALTinterwordstretchfactor}{4}
\providecommand{\BIBentryALTinterwordspacing}{\spaceskip=\fontdimen2\font plus
\BIBentryALTinterwordstretchfactor\fontdimen3\font minus
  \fontdimen4\font\relax}
\providecommand{\BIBforeignlanguage}[2]{{%
\expandafter\ifx\csname l@#1\endcsname\relax
\typeout{** WARNING: IEEEtran.bst: No hyphenation pattern has been}%
\typeout{** loaded for the language `#1'. Using the pattern for}%
\typeout{** the default language instead.}%
\else
\language=\csname l@#1\endcsname
\fi
#2}}
\providecommand{\BIBdecl}{\relax}
\BIBdecl

\bibitem{guo2016ms}
Y.~Guo, L.~Zhang, Y.~Hu, X.~He, and J.~Gao, ``Ms-celeb-1m: A dataset and
  benchmark for large-scale face recognition,'' in \emph{European Conference on
  Computer Vision (ECCV)}, 2016.

\bibitem{lucey2010extended}
P.~Lucey, J.~F. Cohn, T.~Kanade, J.~Saragih, Z.~Ambadar, and I.~Matthews, ``The
  extended cohn-kanade dataset (ck+): A complete dataset for action unit and
  emotion-specified expression,'' in \emph{IEEE Conference on Computer Vision
  and Pattern Recognition (CVPR) Workshop}, 2010.

\bibitem{7438833_tnnls2016}
H.~Lai, P.~Yan, X.~Shu, Y.~Wei, and S.~Yan, ``Instance-aware hashing for
  multi-label image retrieval,'' \emph{IEEE Transactions on Image Processing},
  2016.

\bibitem{9463398_tnnls2021}
X.~Zhang, Y.~Wei, Z.~Li, C.~Yan, and Y.~Yang, ``Rich embedding features for
  one-shot semantic segmentation,'' \emph{IEEE Transactions on Neural Networks
  and Learning Systems}, 2021.

\bibitem{fan2021unsupervised_tcyb2021}
H.~Fan, P.~Liu, M.~Xu, and Y.~Yang, ``Unsupervised visual representation
  learning via dual-level progressive similar instance selection,'' \emph{IEEE
  Transactions on Cybernetics}, 2021.

\bibitem{meng2017identity}
Z.~Meng, P.~Liu, J.~Cai, S.~Han, and Y.~Tong, ``Identity-aware convolutional
  neural network for facial expression recognition,'' in \emph{International
  Conference on Automatic Face and Gesture Recognition (FG)}, 2017.

\bibitem{Zhang_2018_CVPR}
F.~Zhang, T.~Zhang, Q.~Mao, and C.~Xu, ``Joint pose and expression modeling for
  facial expression recognition,'' in \emph{IEEE Conference on Computer Vision
  and Pattern Recognition (CVPR)}, 2018.

\bibitem{yang2018facial}
H.~Yang, U.~Ciftci, and L.~Yin, ``Facial expression recognition by
  de-expression residue learning,'' in \emph{IEEE Conference on Computer Vision
  and Pattern Recognition (CVPR)}, 2018.

\bibitem{8767026_tnnls2020}
D.~Liu, N.~Bellotto, and S.~Yue, ``Deep spiking neural network for video-based
  disguise face recognition based on dynamic facial movements,'' \emph{IEEE
  Transactions on Neural Networks and Learning Systems}, 2020.

\bibitem{liu2021point_tcyb2021}
P.~Liu, Y.~Lin, Z.~Meng, L.~Lu, W.~Deng, J.~T. Zhou, and Y.~Yang, ``Point
  adversarial self-mining: A simple method for facial expression recognition,''
  \emph{IEEE Transactions on Cybernetics}, 2021.

\bibitem{zhang2021pro}
Y.~Zhang, X.~Yu, X.~Lu, and P.~Liu, ``Pro-uigan: Progressive face hallucination
  from occluded thumbnails,'' \emph{arXiv preprint arXiv:2108.00602}, 2021.

\bibitem{zhang2021face_tip2021}
Y.~Zhang, I.~W. Tsang, J.~Li, P.~Liu, X.~Lu, and X.~Yu, ``Face hallucination
  with finishing touches,'' \emph{IEEE Transactions on Image Processing}, 2021.

\bibitem{yan2016image_tmm2016}
Y.~Yan, F.~Nie, W.~Li, C.~Gao, Y.~Yang, and D.~Xu, ``Image classification by
  cross-media active learning with privileged information,'' \emph{IEEE
  Transactions on Multimedia}, 2016.

\bibitem{santander2021pitfalls}
M.~R. Santander, J.~H. Albarrac{\'\i}n, and A.~R. Rivera, ``On the pitfalls of
  learning with limited data: A facial expression recognition case study,''
  \emph{arXiv preprint arXiv:2104.02653}, 2021.

\bibitem{9425436_tnnls2021}
F.~Ma, Y.~Wu, X.~Yu, and Y.~Yang, ``Learning with noisy labels via
  self-reweighting from class centroids,'' \emph{IEEE Transactions on Neural
  Networks and Learning Systems}, 2021.

\bibitem{Li2020}
S.~Li and W.~Deng, ``{A Deeper Look at Facial Expression Dataset Bias},''
  \emph{IEEE Transactions on Affective Computing}, 2020.

\bibitem{erhan2010does_icml2010}
D.~Erhan, A.~Courville, Y.~Bengio, and P.~Vincent, ``Why does unsupervised
  pre-training help deep learning?'' in \emph{ICML}, 2010.

\bibitem{wang2021dense_cvpr2021}
X.~Wang, R.~Zhang, C.~Shen, T.~Kong, and L.~Li, ``Dense contrastive learning
  for self-supervised visual pre-training,'' in \emph{CVPR}, 2021.

\bibitem{wang2018dataset}
T.~Wang, J.-Y. Zhu, A.~Torralba, and A.~A. Efros, ``Dataset distillation,''
  \emph{arXiv preprint arXiv:1811.10959}, 2018.

\bibitem{Liu2014}
P.~Liu, S.~Han, Z.~Meng, and Y.~Tong, ``{Facial expression recognition via a
  boosted deep belief network},'' in \emph{IEEE Conference on Computer Vision
  and Pattern Recognition (CVPR)}, 2014.

\bibitem{mollahosseini2016going}
A.~Mollahosseini, D.~Chan, and M.~H. Mahoor, ``Going deeper in facial
  expression recognition using deep neural networks,'' in \emph{IEEE Winter
  Conference on Applications of Computer Vision (WACV)}, 2016.

\bibitem{li2018occlusion}
Y.~Li, J.~Zeng, S.~Shan, and X.~Chen, ``Occlusion aware facial expression
  recognition using cnn with attention mechanism,'' \emph{IEEE Transactions on
  Image Processing}, 2018.

\bibitem{zhang1998comparison}
Z.~Zhang, M.~Lyons, M.~Schuster, and S.~Akamatsu, ``Comparison between
  geometry-based and gabor-wavelets-based facial expression recognition using
  multi-layer perceptron,'' in \emph{International Conference on Automatic Face
  and Gesture Recognition (FG)}, 1998.

\bibitem{zhang2005active}
Y.~Zhang and Q.~Ji, ``Active and dynamic information fusion for facial
  expression understanding from image sequences,'' \emph{IEEE Transactions on
  Pattern Analysis and Machine Intelligence}, 2005.

\bibitem{tian2002evaluation}
Y.-l. Tian, T.~Kanade, and J.~F. Cohn, ``Evaluation of gabor-wavelet-based
  facial action unit recognition in image sequences of increasing complexity,''
  in \emph{International Conference on Automatic Face and Gesture Recognition
  (FG)}, 2002.

\bibitem{eckhardt2009towards}
M.~Eckhardt, I.~Fasel, and J.~Movellan, ``Towards practical facial feature
  detection,'' \emph{International Journal of Pattern Recognition and
  Artificial Intelligence}, 2009.

\bibitem{yang2007boosting}
P.~Yang, Q.~Liu, and D.~N. Metaxas, ``Boosting coded dynamic features for
  facial action units and facial expression recognition,'' in \emph{IEEE
  Conference on Computer Vision and Pattern Recognition (CVPR)}, 2007.

\bibitem{hu2008multi}
Y.~Hu, Z.~Zeng, L.~Yin, X.~Wei, X.~Zhou, and T.~S. Huang, ``Multi-view facial
  expression recognition,'' in \emph{International Conference on Automatic Face
  and Gesture Recognition (FG)}, 2008.

\bibitem{dahmane2011emotion}
M.~Dahmane and J.~Meunier, ``Emotion recognition using dynamic grid-based hog
  features,'' in \emph{International Conference on Automatic Face and Gesture
  Recognition (FG)}, 2011.

\bibitem{senechal2011combining}
T.~Senechal, V.~Rapp, H.~Salam, R.~Seguier, K.~Bailly, and L.~Prevost,
  ``Combining aam coefficients with lgbp histograms in the multi-kernel svm
  framework to detect facial action units,'' in \emph{International Conference
  on Automatic Face and Gesture Recognition (FG)}, 2011.

\bibitem{valstar2012meta}
M.~F. Valstar, M.~Mehu, B.~Jiang, M.~Pantic, and K.~Scherer, ``Meta-analysis of
  the first facial expression recognition challenge,'' \emph{IEEE Transactions
  on Systems, Man, and Cybernetics, Part B (Cybernetics)}, 2012.

\bibitem{zafeiriou2010sparse}
S.~Zafeiriou and M.~Petrou, ``Sparse representations for facial expressions
  recognition via l 1 optimization,'' in \emph{IEEE Conference on Computer
  Vision and Pattern Recognition (CVPR) Workshop}, 2010.

\bibitem{ying2010facial}
Z.-L. Ying, Z.-W. Wang, and M.-W. Huang, ``Facial expression recognition based
  on fusion of sparse representation,'' in \emph{International Conference on
  Intelligent Computing}, 2010.

\bibitem{liu2013improving}
P.~Liu, S.~Han, and Y.~Tong, ``Improving facial expression analysis using
  histograms of log-transformed nonnegative sparse representation with a
  spatial pyramid structure,'' in \emph{International Conference on Automatic
  Face and Gesture Recognition (FG)}, 2013.

\bibitem{zhong2012learning}
L.~Zhong, Q.~Liu, P.~Yang, B.~Liu, J.~Huang, and D.~N. Metaxas, ``Learning
  active facial patches for expression analysis,'' in \emph{IEEE Conference on
  Computer Vision and Pattern Recognition (CVPR)}, 2012.

\bibitem{8101548}
X.~{Xiang} and T.~D. {Tran}, ``Linear disentangled representation learning for
  facial actions,'' \emph{IEEE Transactions on Circuits and Systems for Video
  Technology}, 2018.

\bibitem{8103056}
S.~{Dai} and H.~{Man}, ``Mixture statistic metric learning for robust human
  action and expression recognition,'' \emph{IEEE Transactions on Circuits and
  Systems for Video Technology}, 2018.

\bibitem{9197663}
S.~{Xie}, H.~{Hu}, and Y.~{Chen}, ``Facial expression recognition with
  two-branch disentangled generative adversarial network,'' \emph{IEEE
  Transactions on Circuits and Systems for Video Technology}, 2020.

\bibitem{Pan2019}
B.~Pan, S.~Wang, and B.~Xia, ``{Occluded facial expression recognition enhanced
  through privileged information},'' in \emph{ACM International Conference on
  Multimedia (MM)}, 2019.

\bibitem{Huang2020}
L.~Shan and D.~Weihong, ``{Deep Facial expression recognition: A survey},''
  \emph{IEEE Transactions on Affective Computing}, 2020.

\bibitem{zhu2005semi}
X.~J. Zhu, ``Semi-supervised learning literature survey,'' Tech. Rep., 2005.

\bibitem{tarvainen2017mean}
A.~Tarvainen and H.~Valpola, ``Mean teachers are better role models:
  Weight-averaged consistency targets improve semi-supervised deep learning
  results,'' in \emph{Annual Conference on Neural Information Processing
  Systems (NeurIPS)}, 2017.

\bibitem{laine2016temporal}
S.~Laine and T.~Aila, ``Temporal ensembling for semi-supervised learning,'' in
  \emph{International Conference on Learning Representations (ICLR)}, 2017.

\bibitem{DBLP:journals/corr/abs-1809-09925}
Y.~Luo, R.~Ji, T.~Guan, J.~Yu, P.~Liu, and Y.~Yang, ``Every node counts:
  Self-ensembling graph convolutional networks for semi-supervised learning,''
  \emph{Pattern Recognition}, 2020.

\bibitem{radosavovic2018data}
I.~Radosavovic, P.~Doll{\'a}r, R.~Girshick, G.~Gkioxari, and K.~He, ``Data
  distillation: Towards omni-supervised learning,'' in \emph{IEEE Conference on
  Computer Vision and Pattern Recognition (CVPR)}, 2018.

\bibitem{cai2018probabilistic}
J.~Cai, Z.~Meng, A.~S. Khan, Z.~Li, J.~O'Reilly, and Y.~Tong, ``Probabilistic
  attribute tree in convolutional neural networks for facial expression
  recognition,'' \emph{arXiv preprint arXiv:1812.07067}, 2018.

\bibitem{johnson2013accelerating_nips2013}
R.~Johnson and T.~Zhang, ``Accelerating stochastic gradient descent using
  predictive variance reduction,'' in \emph{Advances in Neural Information
  Processing Systems (NeurIPS)}, 2013.

\bibitem{Luo_2019_CVPR}
Y.~Luo, L.~Zheng, T.~Guan, J.~Yu, and Y.~Yang, ``Taking a closer look at domain
  shift: Category-level adversaries for semantics consistent domain
  adaptation,'' in \emph{IEEE Conference on Computer Vision and Pattern
  Recognition (CVPR)}, 2019.

\bibitem{zhang2019category}
Q.~Zhang, J.~Zhang, W.~Liu, and D.~Tao, ``Category anchor-guided unsupervised
  domain adaptation for semantic segmentation,'' in \emph{Annual Conference on
  Neural Information Processing Systems (NeurIPS)}, 2019.

\bibitem{li2018reliable}
S.~Li and W.~Deng, ``Reliable crowdsourcing and deep locality-preserving
  learning for unconstrained facial expression recognition,'' \emph{IEEE
  Transactions on Image Processing}, 2018.

\bibitem{goodfellow2015challenges}
I.~J. Goodfellow, D.~Erhan, P.~L. Carrier, A.~Courville, M.~Mirza, B.~Hamner,
  W.~Cukierski, Y.~Tang, D.~Thaler, D.-H. Lee \emph{et~al.}, ``Challenges in
  representation learning: A report on three machine learning contests,''
  \emph{Neural Networks}, 2015.

\bibitem{lyons1998japanese}
M.~J. Lyons, S.~Akamatsu, M.~Kamachi, J.~Gyoba, and J.~Budynek, ``The japanese
  female facial expression (jaffe) database,'' in \emph{International
  Conference on Automatic Face and Gesture Recognition (FG)}, 1998.

\bibitem{pantic2005web}
M.~Pantic, M.~Valstar, R.~Rademaker, and L.~Maat, ``Web-based database for
  facial expression analysis,'' in \emph{IEEE International Conference on
  Multimedia and Expo (ICME)}, 2005.

\bibitem{7553523}
K.~Zhang, Z.~Zhang, Z.~Li, and Y.~Qiao, ``Joint face detection and alignment
  using multitask cascaded convolutional networks,'' \emph{IEEE Signal
  Processing Letters}, 2016.

\bibitem{Parkhi15}
O.~M. Parkhi, A.~Vedaldi, and A.~Zisserman, ``Deep face recognition,'' in
  \emph{{British Machine Vision Conference (BMVC)}}, 2015.

\bibitem{he2016deep}
K.~He, X.~Zhang, S.~Ren, and J.~Sun, ``Deep residual learning for image
  recognition,'' in \emph{IEEE Conference on Computer Vision and Pattern
  Recognition (CVPR)}, 2016.

\bibitem{NEURIPS2019_9015}
A.~Paszke, S.~Gross, F.~Massa, A.~Lerer, J.~Bradbury, G.~Chanan, T.~Killeen,
  Z.~Lin, N.~Gimelshein, L.~Antiga, A.~Desmaison, A.~Kopf, E.~Yang, Z.~DeVito,
  M.~Raison, A.~Tejani, S.~Chilamkurthy, B.~Steiner, L.~Fang, J.~Bai, and
  S.~Chintala, ``Pytorch: An imperative style, high-performance deep learning
  library,'' in \emph{Annual Conference on Neural Information Processing
  Systems (NeurIPS)}, 2019.

\bibitem{li2017reliable}
S.~Li, W.~Deng, and J.~Du, ``Reliable crowdsourcing and deep
  locality-preserving learning for expression recognition in the wild,'' in
  \emph{IEEE Conference on Computer Vision and Pattern Recognition (CVPR)},
  2017.

\bibitem{luo2019significance}
Y.~Luo, P.~Liu, T.~Guan, J.~Yu, and Y.~Yang, ``Significance-aware information
  bottleneck for domain adaptive semantic segmentation,'' in
  \emph{International Conference on Computer Vision (ICCV)}, 2019.

\bibitem{luo2020adversarial_nips2020}
------, ``Adversarial style mining for one-shot unsupervised domain
  adaptation,'' in \emph{NeurIPS}, 2020.

\bibitem{luo2021category_pami2021}
Y.~Luo, P.~Liu, L.~Zheng, T.~Guan, J.~Yu, and Y.~Yang, ``Category-level
  adversarial adaptation for semantic segmentation using purified features,''
  \emph{IEEE Transactions on Pattern Analysis and Machine Intelligence}, 2021.

\bibitem{tjio2021adversarial_wacv2021}
G.~Tjio, P.~Liu, J.~T. Zhou, and R.~S.~M. Goh, ``Adversarial semantic
  hallucination for domain generalized semantic segmentation,'' in \emph{WACV},
  2021.

\bibitem{zhao2018feature}
S.~Zhao, H.~Cai, H.~Liu, J.~Zhang, and S.~Chen, ``Feature selection mechanism
  in cnns for facial expression recognition.'' in \emph{{British Machine Vision
  Conference (BMVC)}}, 2018.

\bibitem{fan2018multi}
Y.~Fan, J.~C. Lam, and V.~O. Li, ``Multi-region ensemble convolutional neural
  network for facial expression recognition,'' in \emph{International
  Conference on Artificial Neural Networks (ICANN)}, 2018.

\bibitem{cai2019improving}
J.~Cai, ``Improving person-independent facial expression recognition using deep
  learning,'' \emph{Ph.D. Thesis}, 2019.

\bibitem{wen2017ensemble}
G.~Wen, Z.~Hou, H.~Li, D.~Li, L.~Jiang, and E.~Xun, ``Ensemble of deep neural
  networks with probability-based fusion for facial expression recognition,''
  \emph{Cognitive Computation}, 2017.

\bibitem{tang2013deep}
Y.~Tang, ``Deep learning using linear support vector machines,'' in
  \emph{International Conference on Machine Learning (ICML) Workshop}, 2013.

\bibitem{breuer2017deep}
R.~Breuer and R.~Kimmel, ``A deep learning perspective on the origin of facial
  expressions,'' \emph{Master Thesis}, 2017.

\end{thebibliography}
